\pdfoutput=1

\documentclass[10pt,twocolumn,letterpaper]{article}

\usepackage[pagenumbers]{wacv} 

\usepackage{graphicx}
\usepackage{amsmath}
\usepackage{amssymb}
\usepackage{booktabs}

\usepackage{comment}
\usepackage[disable,textsize=tiny]{todonotes}
\usepackage{mathtools}
\usepackage{amsthm}
\usepackage{graphicx}
\usepackage[normalem]{ulem}
\useunder{\uline}{\ul}{}
\usepackage{pifont}
\usepackage{wrapfig}
\usepackage{caption}
\usepackage{subcaption}
\usepackage{wrapfig}

\usepackage[pagebackref,breaklinks,colorlinks]{hyperref}

\usepackage[capitalize]{cleveref}
\crefname{section}{Sec.}{Secs.}
\Crefname{section}{Section}{Sections}
\Crefname{table}{Table}{Tables}
\crefname{table}{Tab.}{Tabs.}

\def\wacvPaperID{598} 
\def\confName{WACV}
\def\confYear{2025}

\begin{document}

\title{Finding Dino: A Plug-and-Play Framework for Zero-Shot Detection of Out-of-Distribution Objects Using Prototypes}

\author{
    Poulami Sinhamahapatra \textsuperscript{1, 2},
    Franziska Schwaiger \textsuperscript{1},
    Shirsha Bose \textsuperscript{1, 2},
    Huiyu Wang \textsuperscript{1},\\
    Karsten Roscher \textsuperscript{1}, and 
    Stephan Günnemann \textsuperscript{2}\\
    \textsuperscript{1} Fraunhofer IKS, Germany \\
    \textsuperscript{2} Technical University of Munich, Germany 
}

\maketitle

\begin{abstract}
  Detecting and localising unknown or out-of-distribution (OOD) objects in any scene can be a challenging task in vision, particularly in safety-critical cases involving autonomous systems like automated vehicles or trains. Supervised anomaly segmentation or open-world object detection models depend on training on exhaustively annotated datasets for every domain and still struggle in distinguishing between background and OOD objects. In this work, we present a plug-and-play framework - PRototype-based OOD detection Without Labels (PROWL). It is an inference-based method that does not require training on the domain dataset and relies on extracting relevant features from self-supervised pre-trained models. PROWL can be easily adapted to detect in-domain objects in any operational design domain (ODD) in a zero-shot manner by specifying a list of known classes from this domain. PROWL, as a first zero-shot unsupervised method, 
  achieves state-of-the-art results on the RoadAnomaly and RoadObstacle datasets provided in road driving benchmarks - SegmentMeIfYouCan (SMIYC) and Fishyscapes, as well as comparable performance against existing supervised methods trained without auxiliary OOD data.
  We also demonstrate its generalisability to other domains such as rail and maritime.
  
\end{abstract}

\vspace{-5mm}
\section{Introduction}
\label{sec:intro}

\begin{figure}[htbp]
    \centering
     \includegraphics[width=\linewidth]{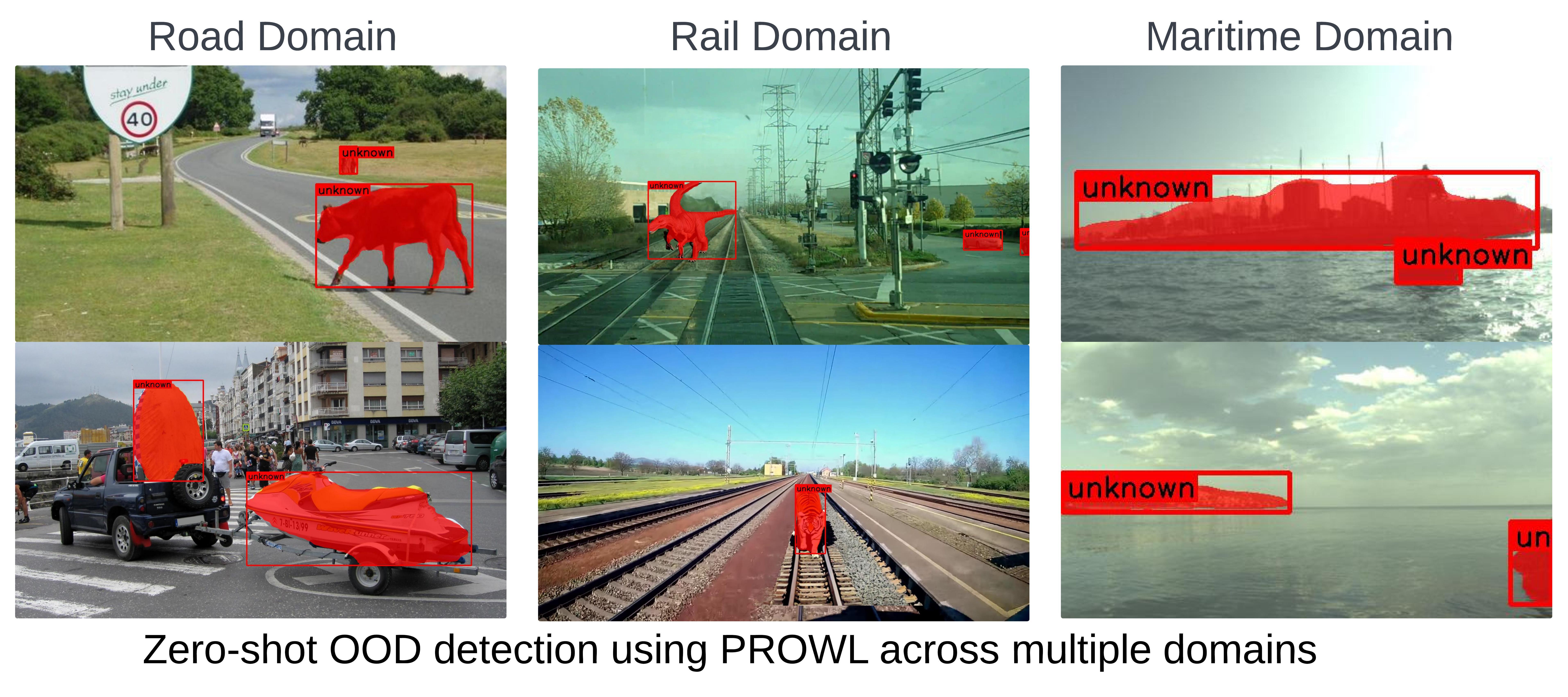}
   \caption{Sample results for zero-shot detection of OOD objects with PROWL across multiple domains: road driving (test set of RoadAnomaly dataset from SMIYC benchmark \cite{chanSegmentMeIfYouCanBenchmarkAnomalya}), rail (created test set with inpainted OOD objects on RailSem19 \cite{zendelRailSem19DatasetSemantic2019}) and maritime scene (test set of marine obstacle detection dataset \cite{bovconMaSTr1325DatasetTraining2019}). Detected OOD objects are marked as `unknown' in red.
   }
   \label{fig:teaser}
   \vspace{-7mm}
\end{figure}
\vspace{-1mm}
Artificial Intelligence (AI) has become a cornerstone of autonomous systems - especially in the perception of the surroundings. Systems operating in the real world must dynamically adapt to any situation in open-world settings. This means that in any given scene: the system should be able to understand its context, usually through the detection and localisation of relevant objects. To this end, AI models are typically trained extensively based on closed-set object categories in the given operational design domain (ODD). With high-quality publicly available datasets, such as Cityscapes \cite{cordtsCityscapesDatasetSemantic2016}, 
RailSem19 \cite{zendelRailSem19DatasetSemantic2019} and MODD \cite{bovconMaSTr1325DatasetTraining2019}, several State-of-the-Art (SOTA) deep neural networks (DNNs) provide outstanding performances for a closed set of object categories. However, they are unable to identify and categorise unknown objects, i.e. objects that do not belong to any of the training classes. One can encounter obstacles that were never learned in the training data such as a random animal on the road or unknown floating obstacles in front of unmanned surface vehicles (USVs) in maritime applications. Due to the open world setting, there can be numerous such unknown or out-of-distribution (OOD) objects at any time and it is almost impossible to train DNNs exhaustively with annotated datasets on all possible known object categories and object variations, especially in complex domains such as autonomous driving.

In contrast to image classification, where OOD detection is a well-defined and widely researched topic, the main challenge in the context of camera-based object detection is the explicit distinction between an unknown object and the common \emph{background}, i.e. anything in the scene that is not relevant. Therefore, most existing approaches rely on supervised anomaly segmentation and often use selected OOD samples during training or fine-tuning as auxiliary OOD data\cite{raiUnmaskingAnomaliesRoadScene,chanEntropyMaximizationMeta2021, ackermannMaskomalyZeroShotMask2023a, tianPixelwiseEnergybiasedAbstention2022, leeSimpleUnifiedFramework2018}. Especially the latter is a severe limitation when trying to detect things that were not known at the training time of a model. Open world object detection \cite{zoharPROBProbabilisticObjectness2023, guptaOWDETROpenworldDetection2022a, fontanelDetectingUnknownObject2022a} on the other hand gained some attention recently but is struggling with the application-dependent understanding what a relevant object is.

In this work, we propose a novel framework for the detection of unknown objects in an image: PRototype-based 
OOD detection Without Labels (PROWL). It can detect an arbitrary number of unknown objects in scenes from any domain in a zero-shot plug-and-play manner, i.e. without any additional supervised model training or fine-tuning on samples from the target domain, as illustrated in Fig. \ref{fig:teaser}.
It leverages the rich and diverse features from \textit{frozen} foundation models such as self-supervised trained DINOv2 \cite{oquabDINOv2LearningRobust} to robustly capture the known object categories as prototypes in a prototype feature bank. The similarity to those prototypes is then used to calculate 
a pixel-level similarity score for each given test image. This score can be thresholded to detect OOD pixels. 
Additionally, we propose to combine foreground masks provided by unsupervised segmentation with those pixel-level scores to refine them into high-quality masks for individual instances of OOD or unknown objects. 
Making use of pre-trained visual features from foundation models which provide robust representations of almost any object, PROWL can easily generalize across application domains by simply specifying the list of ODD classes to determine the corresponding prototype feature bank. It can detect unknown objects without additional model training in just simple inference steps and performs comparably to existing supervised methods. To the best of our knowledge, \textit{PROWL is the first end-to-end framework for zero-shot unsupervised OOD object detection in any scene} without any explicit training on ground truth (GT) classes. Within the framework of PROWL, we compare the performance of different unsupervised segmentation and detection methods based on the quality of foreground masks generated. In the absence of a direct baseline for zero-shot anomaly segmentation in a multi-object scene, we further compare our results with other supervised segmentation methods from the SMIYC~\cite{chanSegmentMeIfYouCanBenchmarkAnomalya} benchmark based on established metrics and datasets. 

In summary, the key contributions of our proposed framework PROWL are:
\begin{itemize}
    \vspace{-2mm}
    \item PROWL is the first zero-shot unsupervised OOD object detection and segmentation framework that can sufficiently and reliably distinguish OOD objects from background.
    \vspace{-2mm}
    \item PROWL relies on frozen features from self-supervised foundation models without need for training or fine-tuning on domain data, by simply creating an offline prototype feature bank using very few object samples per class.
    \vspace{-2mm}
    \item PROWL can be applied as plug-and-play module adaptable  to any scene in a new domain without domain-specific training. We demonstrate this by applying to new domains beyond road driving such as rail and maritime domains. For the rail domain, we additionally demonstrate by creating a sample dataset with in-painted OOD objects.
    \vspace{-2mm}
    \item PROWL outperforms fully supervised methods trained without auxiliary OOD data given in road driving benchmark SMIYC \cite{chanSegmentMeIfYouCanBenchmarkAnomalya} on RoadObstacle datasets, as well as show comparable performance with other supervised methods in RoadAnomaly and Fishyscapes \cite{blumFishyscapesBenchmarkMeasuring2021} benchmark.
    
\end{itemize}

\vspace{-2mm}
\section{Related Work}
\vspace{-1mm}
\label{sec:related_work}
In vision tasks, detecting the OOD object has been formulated under different banners.


\textbf{Open World Object Detection (OWOD):}
As compared to standard closed-world object detection, OWOD poses several challenges such as generating quality candidate proposals on potentially unknown objects or distinguishing the unknown objects from the background. Recent methods \cite{zoharPROBProbabilisticObjectness2023, guptaOWDETROpenworldDetection2022a, fontanelDetectingUnknownObject2022a} have explored probabilistic models based on objectness score of the unknown object and learning novel classes via incremental learning and retraining. But they still have much room for improvement in distinguishing the unknown class from the background.

\textbf{Anomaly Detection:}
Anomaly or out-of-distribution (OOD) detection was initially conducted in the context of image classification. OOD detection has been widely used for finding deviations from in-distribution (ID), i.e. training data. It encompasses both deviations in distribution shift such as perturbations, weather, or lighting conditions as well as changes in semantic classes unseen during training. Methods originating from image classification focused on developing techniques that aimed to quantify uncertainty in confidence values produced by classification outputs of DNNs (e.g., Maximum Softmax Probability \cite{hendrycksBaselineDetectingMisclassified2018}, Mahalanobis distance \cite{leeSimpleUnifiedFramework2018}). Other methods find anomalies by estimating the likelihood with generative models \cite{renLikelihoodRatiosOutofDistribution2019} or training discriminatively with negative or auxiliary data from OOD samples like ODIN \cite{liangEnhancingReliabilityOutofdistribution2020}, Outlier exposure \cite{hendrycksDeepAnomalyDetection2019}. Another line of work is finding anomalies or defective parts in an object, such as defective part detection in Industrial Anomaly Detection \cite{liuDeepIndustrialImage2024}. Although developed as image-level anomaly detection, most of these methods can be applied to anomaly segmentation by finding potential anomalies based on the confidences of each pixel.

\begin{figure*}[!h]
    \vspace{-3mm}
    \centering
     \includegraphics[width=0.85\textwidth]{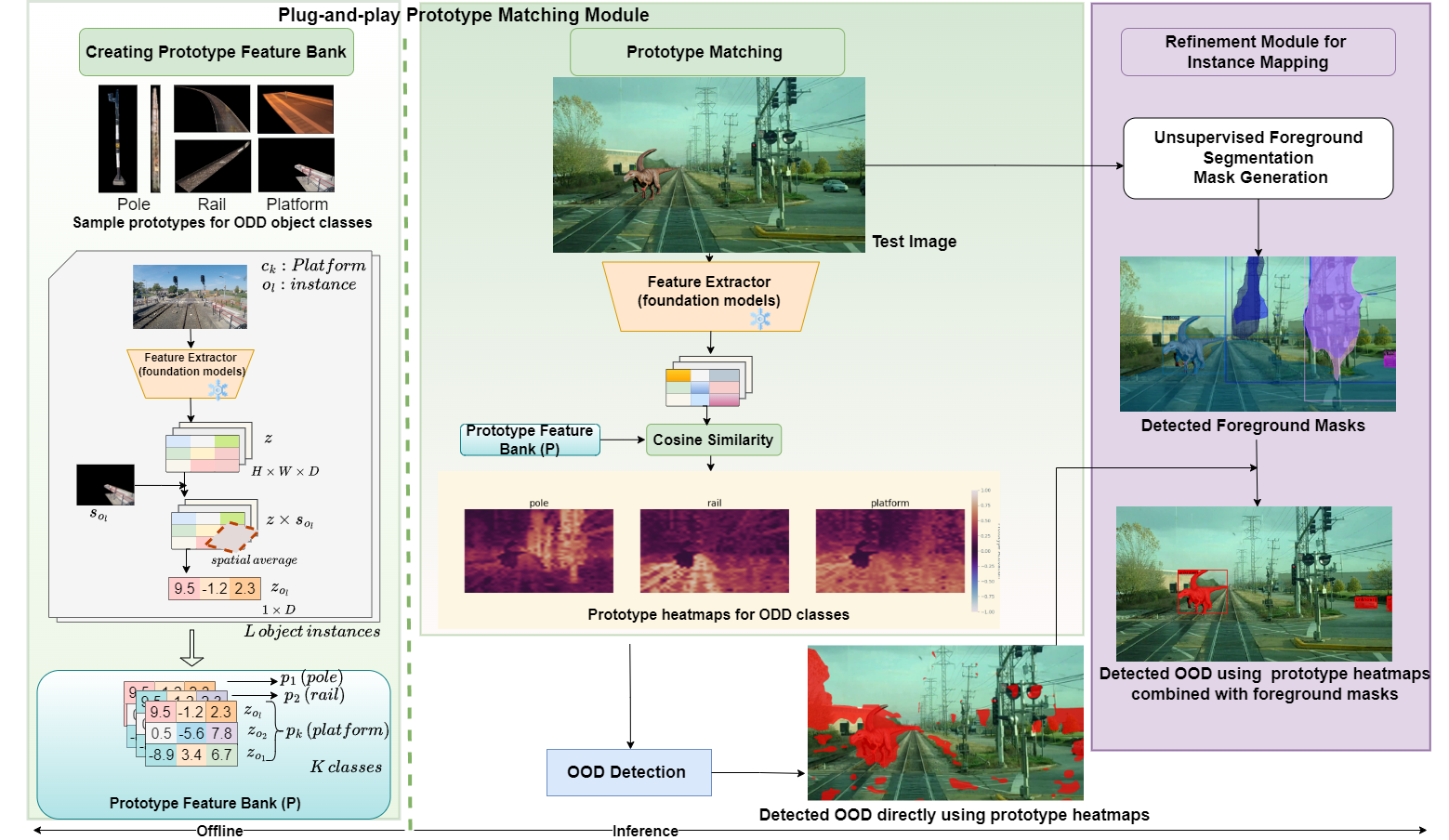}
   \caption{Overview of our proposed framework PROWL. Firstly, in the plug-and-play prototype matching module, prototype feature bank is created by extracting features from pre-trained foundation models corresponding to few segmented object samples for specified list of ODD object classes. Using this feature bank, prototype  matching is performed for the given test image to generate corresponding heatmaps for each object class. Heatmaps show maximum activation (in yellow) wherever the given object is found in test image. In the OOD detection step, the object pixels not satisfying given similarity thresholds are detected as OOD or `unknown' (in red). For less noisy and precise OOD detection, we combine an additional refinement step with prototype heatmaps where foreground masks for every objects in the scene are first extracted in an unsupervised manner. Finally, these foreground masks are detected as either an ODD class or an OOD.
  }
   \label{fig:architecture}
   \vspace{-5mm}
\end{figure*}

\textbf{Anomaly Segmentation:} 
The goal is to predict anomaly probabilities for each pixel in an image. Different works use discriminative or generative approaches \cite{dibiasePixelwiseAnomalyDetection2021, lisDetectingUnexpectedImage2019}. Most methods rely on auxiliary OOD data during training. For example, Max Entropy \cite{chanEntropyMaximizationMeta2021} predicts high entropy in anomalous regions and reduces false positives using a meta-classifier on OOD data. DenseHybrid \cite{grcicDenseHybridHybridAnomaly2022} combines discriminative and generative modeling. However, pixel-based reasoning often produces noisy anomaly scores, especially for border pixels and poorly localized anomalies. Recent approaches focus on mask-based methods that capture anomalies as whole objects. These methods predict regions instead of pixels, resulting in fewer false predictions \cite{chengMaskedattentionMaskTransformer2022}. RbA \cite{nayalRbASegmentingUnknown2023}, EAM \cite{grcicAdvantagesMasklevelRecognition2023}, Maskomaly \cite{ackermannMaskomalyZeroShotMask2023a}, Mask2anomaly \cite{raiUnmaskingAnomaliesRoadScene} and S2M \cite{zhao2024segment}
utilize mask-based classification. However, all these methods are trained with supervised labels, including additional synthetic data \cite{zhao2024segment} and even sometimes \textit{exposed to OOD data (auxiliary data)}. Our proposed approach learns every possible object in the scene as foreground masks without the notion of OOD/ODD object class and performs OOD object detection separately. It eliminates the need for confidence based anomaly scores from supervised training to discover OOD objects.

\vspace{-2mm}
\section{Method}

\label{sec:method}
\vspace{-2mm}
In this section, we provide an overview of our framework PROWL. As illustrated in the architecture diagram in Fig. \ref{fig:architecture}, PROWL comprises of three modules: a plug-and-play prototype matching module (Sec. \ref{sec:prototype_matching}) followed by OOD detection module (Sec. \ref{sec:ood_detection}), and lastly a refinement module used to generate foreground masks for enhanced OOD detection (Sec. \ref{sec:fg_mask_gen}).

\vspace{-2mm}
\subsection{Plug-and-play Prototype Matching Module}

\label{sec:prototype_matching}
\vspace{-2mm}
The first step in our pipeline is to create the feature bank with prototype features for every object class specified in the ODD. Subsequently, every pixel is assigned a class via prototype matching step.

 \textbf{Creating the prototype feature bank:} We aim to create an \textit{offline} `prototype feature bank' which consists of global feature space representations corresponding to domain object classes. 
 Pre-trained features from foundation models like DINO \cite{caronEmergingPropertiesSelfSupervised2021} use knowledge distillation via a teacher-student network for learning in a self-supervised approach. Authors\cite{caronEmergingPropertiesSelfSupervised2021}  observe that a self-supervised Vision Transformer (ViT) can learn to a great extent the underlying perceptual grouping of image patches and semantic correspondences across images and image domains. This property is even strengthened for DINOv2 features \cite{oquabDINOv2LearningRobust} which are trained on much larger image corpora and distilled to smaller models. Thus, we utilize the robust general-purpose visual frozen features from such feature extractors for object classes in ODD using a minimal number of samples from the train split. We assume a list of ODD object classes (expert-specified) that one can expect in a given scene, say $K$ known classes $C = \{ c_1, c_2,...,c_K \} $ and a list of prototype vectors for each class 
 $ P = \{ p_1, p_2,...,p_K \} $. Let us assume $L$ object instances contribute to each prototype vector $ p_k$ for each class $c_k$. Let the output of the frozen feature extractor ($g$) be $z = g(x)$ for an image $x$. Assuming $D, h, w$ as embedding dimension, token height and width respectively for given backbone, size of $z$ is $1 \times D \times h \times w$. For each object instance $o_l \in c_k$ class, the GT mask $ s_{o_l} $ is multiplied as a binary mask with $z$ to extract instance-specific features. Finally, a spatial averaging on last 2 dimensions is performed to obtain $1D$ feature representation of a  prototype instance, given as: 

\vspace{-3mm}
 \begin{equation}
    \vspace{-2mm}
     z_{o_l} = \mbox{mean}(z * s_{o_l})
 \end{equation}
 
 The prototype vector list $p_k$ is extended with $1D$ vectors $z_{o_l}$
 until $L$ object instances of $c_k$ class is added, repeating for all the $K$ ODD object classes. Depending on the complexity of the dataset, only few prototype samples can suffice $L \in \{5, 20\}$ for each object (Sec \ref{sec:results_ablation}) as compared to supervised training methods which require lots of training samples.

 \textbf{Prototype matching:} For each test image $x$, the inference output of the feature extractor is given as $z$. Using prototype feature bank $P$ for K classes, K prototype heatmaps are calculated based on maximum cosine similarity between the prototype vector list ${p_k}$ 
 and $z$ 
 , for respective class $c_k$  given as:
 

\vspace{-4mm}
\begin{equation}
    \vspace{-2mm}
    \label{eq:cosine_similarity}
    h_k =  \mbox{max} (z \cdot p_k )
\end{equation}

\noindent
Taking maximum over L instances, heatmaps $h_k$ of size $h \times w$  are obtained which are then upsampled to image resolution.
The list of prototype heatmaps for all the $K$ ODD classes is given as, $ H = \{ h_1, h_2,...,h_K \}$. For each pixel $[i, j]$ in $x$, the assigned class label ($y$) and score ($v$) is given as: 

\vspace{-3mm}
\begin{align}
    \vspace{-2mm}
    \label{eq:pixel_classify}
    y_{[i,j]} & = \mbox{argmax}(H)
\end{align} 

\vspace{-3mm}
\begin{align}
    \vspace{-2mm}
    \label{eq:pixel_score}
    v_{[i,j]} & = \mbox{max}(H) 
\end{align} 



\vspace{-1mm}
\subsection{OOD Detection}
\vspace{-1mm}
\label{sec:ood_detection}

The next step is to detect the OOD pixels following the prototype-based classification of every pixel in Eq \ref{eq:pixel_classify}. This is done by comparing the cosine similarity scores obtained in Eq \ref{eq:pixel_score} with a given threshold $t, t \in [0, 1] $. For every pixel, we calculate an \textit{inverse normalised cosine similarity} (INCS) score as:
\vspace{-2mm}
\begin{align}
    \label{eq:inverse_pixel_score}
    w_{[i,j]} & = 1 - \mbox{norm}(v_{[i,j]}),  \mbox{norm}(a) = \frac{a - \mbox{min}(a)}{\mbox{max}(a)- \mbox{min}(a)}
\end{align} 

Thus, pixels where $ w_{[i,j]} > t$ is designated as an OOD pixel, otherwise retains the class label $y_{[i,j]}$. 
This per-pixel OOD detection based on prototype heatmaps is usually found to be quite reliable, however it could sometimes show noisy detections when the OOD pixels do not belong to a relevant object.
Thus, the output of PROWL can be further refined by introducing instance-level foreground masks given in Sec. \ref{sec:fg_mask_gen} in combination with the prototype heatmaps.

\vspace{-1mm}
\subsection{Refinement module using Foreground Masks}
\vspace{-2mm}
\label{sec:fg_mask_gen}


In the refinement module, we focus on the generation of foreground masks for every object in the image irrespective of their classes. SOTA unsupervised segmentation methods such as STEGO or CutLER can provide such masks with high quality. Let the foreground masks generated using these methods be denoted as $M = \{ m_1, m_2,...,m_N \}$. STEGO\cite{hamiltonUNSUPERVISEDSEMANTICSEGMENTATION2022a} is a semantic segmentation model that distills pre-trained unsupervised visual features from DINO \cite{caronEmergingPropertiesSelfSupervised2021} into semantic clusters using contrastive loss, thus discovering and segmenting semantic objects without human supervision for each dataset. CutLER \cite{wangCutLearnUnsupervised2023a} is an approach for training unsupervised object detection and segmentation models. It is trained exclusively on unlabeled ImageNet \cite{dengImageNetLargescaleHierarchical2009} data without any additional in-domain data. 
It uses \textit{MaskCut} strategy to discover multiple coarse object masks, which is used to train a detector through several rounds of self-training to detect multiple foreground objects and corresponding instance segmentation masks. Contrary to STEGO, CutLER does not provide dense segmentation as output, rather it provides a zero-shot object detection and instance segmentation for detected foreground objects. 

Since these models were trained on huge generic datasets in a self-supervised manner, they can reliably detect multiple foreground object masks in a scene without the notion of ODD/OOD object class. These masks tend to capture the objectness of every entity in the scene without learning it as background as in supervised detection cases. Thus, every mask $m$ where the majority of pixels is designated as OOD (in Sec. \ref{sec:ood_detection}) is now considered to be OOD for the entire mask. As shown in Figure \ref{fig:architecture}, the OOD objects \textit{dinosaur} and \textit{passenger car} were correctly detected using prototype heatmaps in PROWL as they did not have high similarity with any of the ODD classes listed in the prototype bank, however, additional pixels were also spuriously detected as OOD. While PROWL in combination with foreground masks, correctly detected and precisely localised the exact OOD object masks as `unknown'. 
\vspace{-2mm}
\section{Implementation Details}
\vspace{-1mm}
\label{sec:implementation}

\subsection{Datasets}
\label{sec:datasets}
\vspace{-2mm}
We demonstrate the plug-and-play performance of our methods by evaluating them in three different ODD:
 
 \textbf{Road driving scene} - We generated prototypes for the urban driving road ODD scene from \textit{Cityscapes} \cite{cordtsCityscapesDatasetSemantic2016}. It is a benchmark suite with both pixel and instance-level semantic scene understanding. 
 For evaluation on datasets with real OOD objects in the road scene, we refer to the anomaly segmentation in \textit{SegmentMeIfYouCan} (SMIYC) \cite{chanSegmentMeIfYouCanBenchmarkAnomalya}. 
 It provides two novel real-world datasets: \textit{RoadAnomaly21(RA)} and \textit{RoadObstacle21(RO)}. RA consists of $100$ test and $10$ validation images with real objects or animals as OOD appearing anywhere in the scene. In contrast, RO has OOD objects (or obstacles) appearing on the road or ego track. SMIYC withholds the GT for the test set, where the scores are only accessible by submitting the method to the official benchmark. Further, we also evaluate on \textit{FS Static} subset of \textit{Fishyscapes} \cite{blumFishyscapesBenchmarkSafe2019} anomaly segmentation benchmark. It consists of $30$ test images with generic objects taken from PASCAL VOC \cite{everinghamPascalVisualObject2010} synthetically overlayed on Cityscapes images. 
 
 \textbf{Rail scene} - For rail ODD, we use \textit{RailSem19}\cite{zendelRailSem19DatasetSemantic2019} dataset which provides diverse images taken from an ego-perspective of a rail vehicle (trains or trams) along with extensive semantic annotations including rails. 
 Regarding OOD detection, there exist no publicly available datasets with OOD objects in the rail scene yet. Thus, we create our in-house in-painted test data 
 as detailed in Suppl. A. 
 
 \textbf{Maritime scene} - Obstacle detection and segmentation in maritime ODD (\textit{MODD}) scenario is prevalent for autonomous operations of unmanned surface vehicles (USVs). Here, the task is to detect and segment all obstacles beyond \textit{sky, sea} classes as `unknown'. We formulate this as an OOD detection task, where prototypes are generated from train split of \textit{MaStr1325} (Maritime Semantic Segmentation Training Dataset) \cite{bovconMaSTr1325DatasetTraining2019}. For OOD detection evaluation, we evaluate on corresponding test dataset. 

\vspace{-2mm}  
\subsection{Experimental Setup}
\label{sec:exp_setup}
\vspace{-1mm}
PROWL is completely based on inference on frozen features from foundation models without using any domain data for training.The chosen feature extractor is ViT-S/14 of DINO-v2\cite{oquabDINOv2LearningRobust} pre-trained on a dataset of 142 million images. 
Using a \textit{held-out validation set} (unused samples from train split), we observed good performance with a \textit{default} INCS threshold (Eq. \ref{eq:inverse_pixel_score}) of $0.55$ and a low CutLER detector threshold of $0.2$. We note that for zero-shot OOD detection keeping a low detector thresholds helps in detecting all possible sizes of object instances. We further report that depending upon the size of the OOD objects and complexity of the OOD datasets, this threshold can be further optimised to suffice even at higher values as shown in the ablation study (Sec. \ref{sec:results_ablation}). 

\vspace{-2mm}
\subsection{Evaluation metric} 
\vspace{-1mm}
To the best of our knowledge, \textit{there exists no prior work, benchmarks and evaluation metrics  on zero-shot inference on foundational models for detection of OOD objects}. Hence, we compare with existing road anomaly segmentation benchmarks which provide OOD datasets and metrics, such as AUPRC and False Positive Rate at True Positive Rate of $95 \%$ (FPR) metrics for pixel-wise evaluation. Further, for evaluation as a binary segmentation task, we select a fixed INCS threshold for all the datasets and predict $1$ (positive) for pixels determined as OOD otherwise $0$. This binary prediction mask is compared with the GT mask and evaluated using Intersection over Union (IoU) for the OOD class and F1 score.


\vspace{-2mm}
\section{Results and Discussion}
\vspace{-2mm}
\label{sec:results}

In this section, different experiments to evaluate the performance of PROWL for OOD detection are shown. In Sec. \ref{sec:results_road}, we present results comparing with State-of-the-Art (SOTA) supervised methods in the road driving scene. 
In Sec. \ref{sec:results_other-odd}, we show the applicability of PROWL into other domains such as rail and maritime scenes where there are no directly comparable benchmarks. Lastly, in Sec \ref{sec:results_ablation}, we provide ablation for different hyper-parameters and strategies used in our implementation.

We report results for our method PROWL in two scenarios: (a) PROWL (Without foreground masks in Sec. \ref{sec:ood_detection})  - OOD pixels are directly obtained by thresholding the INCS scores from prototype heatmaps, (b) PROWL + refinement module using foreground masks (Sec. \ref{sec:fg_mask_gen}) - The foreground masks generated by unsupervised segmentation methods (STEGO \cite{hamiltonUNSUPERVISEDSEMANTICSEGMENTATION2022a}, CutLER \cite{wangCutLearnUnsupervised2023a} as applicable) are combined with prototype heatmaps, such that individual masks (not pixels) are detected as OOD objects or not.

\begin{table}[!h]
\resizebox{\linewidth}{!}{%
\begin{tabular}{lccllllll}
\hline
 &
  \multicolumn{1}{l}{} &
  \multicolumn{1}{l}{} &
  \multicolumn{2}{c}{RoadAnomaly} &
  \multicolumn{2}{c}{FS Static} &
  \multicolumn{2}{c}{RoadObstacle} \\ \cline{4-9} 
Method &
  \multicolumn{1}{l}{Seg. type} &
  \multicolumn{1}{l}{Aux. data} &
  \multicolumn{1}{c}{AUPR} &
  FPR &
  AUPR &
  FPR &
  AUPR &
  FPR \\ \hline
\textit{\textbf{Supervised}} &
  \multicolumn{1}{l}{} &
  \multicolumn{1}{l}{} &
   &
   &
   &
   &
   &
   \\
PEBAL &
  Pixel &
  \ding{51}  &
  45.1 &
  44.6 &
  \textbf{92.1} &
  \textbf{1.5} &
  - &
   \\
Max. Entropy &
  Pixel &
  \ding{51}  &
  79.7 &
  19.3 &
  76.3 &
  7.1 &
  - &
  - \\
DenseHybrid &
  Pixel &
  \ding{51}  &
  63.9 &
  43.2 &
  60.0 &
  4.9 &
  - &
  - \\
M2A &
  Mask &
  \ding{51}  &
  79.7 &
  13.5 &
  - &
  - &
  - &
  - \\
RbA &
  Mask &
  \ding{51} &
  \textbf{85.42} &
  \textbf{6.92} &
   &
   &
   &
   \\
EAM &
  Mask &
  \ding{55}  &
  66.7 &
  13.4 &
  {\ul 87.3} &
  {\ul 2.1} &
  - &
  - \\
cDNP &
  Pixel &
  \ding{55} &
  {\ul 79.78} &
  18.18 &
  - &
  - &
  - &
  - \\
Maskomaly &
  Mask &
  \ding{55} &
  70.9 &
  {\ul 11.9} &
  69.5 &
  14.4 &
  0.96* &
  92.10* \\\hline
 \textit{\textbf{\begin{tabular}[c]{@{}l@{}}Zero-shot inference on \\ Foundation models\end{tabular}}} &
  \multicolumn{1}{l}{} &
  \multicolumn{1}{l}{} &
   &
   &
   &
   &
   &
   \\
PROWL &
  Pixel &
  \ding{55} &
  45.98 &
  26.58 &
  {\ul 61.08} &
  {\ul 20.28} &
  11.10 &
  46.97 \\
PROWL + STEGO &
  Mask &
  \ding{55} &
  {\ul 53.97} &
  {\ul 13.52} &
  45.6 &
  26.25 &
  {\ul 40.15} &
  {\ul 14.11} \\
PROWL + CutLER &
  Mask &
  \ding{55} &
  \textbf{75.25} &
  \textbf{1.75} &
  \textbf{70.27} &
  \textbf{8.21} &
  \textbf{73.53} &
  \textbf{5.58} \\ \hline
\end{tabular}%
}
\caption{Results on RoadAnomaly, FS Static and RoadObstacles. We separate the methods based on supervision needed during training. The best results are marked in bold, and second best are underlined in each category. * indicates reproduced results with official code and checkpoints. PROWL+CutLER outperforms zero-shot variants in all datasets as well as supervised methods trained without auxiliary data on RoadObstacle.}
\label{tab:smiyc-val}
\vspace{-4mm}
\end{table}

\begin{figure*}[htbp]
    \centering
     \vspace{-3mm}
     \includegraphics[width=0.60\linewidth]{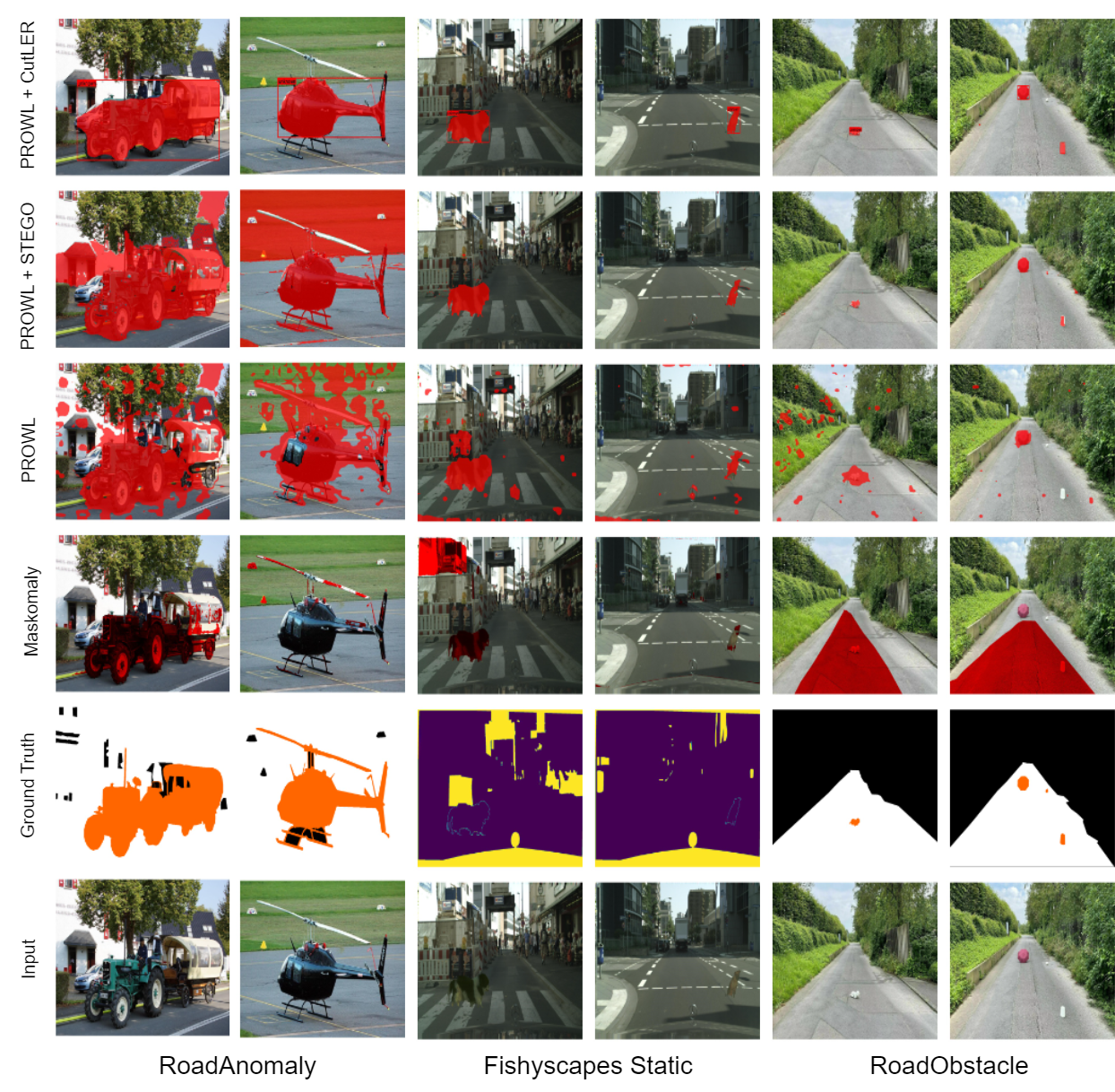}
     \vspace{-3mm}

   \caption{Qualitative comparison of different zero-shot methods with PROWL as compared to supervised baseline Maskomaly. Results generated using fixed thresholds for the RoadAnomaly, FS Static and RoadObstacles datasets. Detected OOD pixel segmentations shown in red. PROWL+CutLER provides qualitatively superior OOD detection and segmentation across all datasets.
   }
   \label{fig:method-comparison}
   \vspace{-5mm}
\end{figure*}

\vspace{-2mm}
\subsection{Comparison with SOTA on road driving scene}
\vspace{-1mm}
\label{sec:results_road}

In the road driving scene, we evaluate on datasets with real OOD objects like RA and RO provided by SMIYC \cite{chanSegmentMeIfYouCanBenchmarkAnomalya} and the synthetic OOD objects in FS-Static given in Fishyscapes \cite{blumFishyscapesBenchmarkMeasuring2021} benchmark. 
We created prototype bank for road scene using $19$ ODD classes from Cityscapes\cite{cordtsCityscapesDatasetSemantic2016} and also followed in SMIYC.

There exists \textit{ no prior work serving as a baseline for zero-shot OOD object detection task} based on foundation models,
it should be noted that all the methods used for comparison with SOTA are fully supervised, i.e. these models are trained with GT instance masks of the $19$ classes of Cityscapes, whereas we rely on few inference steps on frozen self-supervised DINOv2 models. 
We note benchmark (SMIYC, Fishyscapes) evaluation code requires logits based on finite number of closed set categories by varying detector thresholds. However, PROWL is based on zero-shot inference on foundation models which can not output logits on fixed classes making \textit{zero-shot methods not directly comparable with  benchmarks designed for supervised methods}. To still compare with SOTA methods, we calculate evaluation metrics using INCS score (Eq. \ref{eq:inverse_pixel_score}) and regard the corresponding threshold as the variable parameter which is different from detector thresholds. Since SMIYC test GT are not available without benchmark submissions, we report quantitative results only on  \textit{RA} and \textit{RO} (where GT given) in  Table \ref{tab:smiyc-val}, \ref{tab:cityscapes-thresh-fix} and visually impressive qualitative results on SMIYC \textit{test set} in Fig. \ref{fig:teaser}, and Fig. 4, 5 in Supplementary.

\vspace{-1mm}
In Table \ref{tab:smiyc-val}, we compare with SOTA methods using threshold-independent evaluation metrics such as AUPR and FPR at the best TPR (or $95 \%$). We include those highly-ranked SOTA methods from SMIYC benchmark which also report results on mentioned datasets. We indicate the segmentation type of each method as pixel or mask-based. Further, many SOTA methods achieve superior results by discriminative training with select OOD samples (\textit{auxiliary data}) during training or exposing to such OOD samples for fine-tuning.
For comparison purposes, we assume \textit{Maskomaly} \cite{ackermannMaskomalyZeroShotMask2023a} as our direct \textit{supervised baseline} as it is also a zero-shot inference based approach like PROWL, depending on pre-trained Mask2Former \cite{chengMaskedattentionMaskTransformer2022} on Cityscapes, provides reproducible code for considered datasets and does not depend on auxiliary OOD data. 
We show that PROWL with CutLER performs best compared to all the other variants amongst the zero-shot methods. In comparison to \textit{supervised methods trained without auxiliary data}, PROWL with CutLER provides overall best FPR and second best to cDNP\cite{10350977} in AUPR on the RA and overall outperforms on RO datasets. 
We note that most supervised methods do not report results on RO and the reproduced results for Maskomaly show poor performance, also shown in Fig. \ref{fig:method-comparison}.
For FS Static, our zero-shot methods performs better than Maskomaly and comparably similar to other supervised methods.

\begin{table}[htbp]
 \vspace{-3mm}
\centering
\resizebox{0.95\linewidth}{!}{%
\begin{tabular}{lllllll}
\hline
              & \multicolumn{2}{c}{RoadAnomaly}             & \multicolumn{2}{c}{FS Static}               & \multicolumn{2}{l}{RoadObstacle}            \\ \cline{2-7} 
Method        & IoU & F1             & IoU & F1             & IoU & F1             \\ \hline
Maskomaly*          & 46.86 & 56.76 & 23.97 & 29.92 & 9.31  & 11.29 \\ \hline
PROWL & 38.46 & 54.68 & 39.43 & 50.75 & 6.79  & 11.21 \\
PROWL + STEGO        & 56.24 & 69.81 & 39.21 & 49.9  & 29.92 & 35.10 \\
PROWL + CutLER & \textbf{75.22}             & \textbf{85.25} & \textbf{64.79}             & \textbf{72.18} & \textbf{49.16}             & \textbf{53.31} \\ \hline
\end{tabular}%
}
\caption{Comparison of performance of our zero-shot methods as compared to supervised baseline based on binary segmentation with fixed thresholds on respective OOD datasets. * indicates reproduced results with official code and checkpoints with a threshold of 0.9.}
\label{tab:cityscapes-thresh-fix}
\vspace{-3mm}
\end{table}

\vspace{-1mm}
In Table \ref{tab:cityscapes-thresh-fix} and Fig. \ref{fig:method-comparison}, we present the comparative performance of our methods against supervised SOTA baseline (Maskomaly \cite{ackermannMaskomalyZeroShotMask2023a}) as a binary segmentation task. For this, the binary GT masks (where OOD pixels are given as $1$) are compared with the binary masks generated from the predicted OOD pixels via different methods. We evaluate mIoU and F1 scores with fixed INCS thresholds. 
The authors of Maskomaly \cite{ackermannMaskomalyZeroShotMask2023a} report results on the validation set by calculating best threshold every image by checking best F1 score using GT. However, for fair comparison we reproduce all the results for Maskomaly with their reported best fixed confidence threshold of $0.9$ for those datasets.  

\vspace{-0.5mm}
In Table \ref{tab:cityscapes-thresh-fix}, we show that PROWL with CutLER outperforms other zero-shot variants as well as supervised baseline in terms of both mIoU and F1 scores with a significant margin. We observe that there's an overall decreasing trend of performance in most methods across OOD datasets from RA, FS-Static to RO. This can be attributed also to the difficulty of the dataset and the size of OOD object within the image.
 We can correlate this with the qualitative results provided in Fig. \ref{fig:method-comparison}. RA dataset provides real OOD objects which are comparably big like the \textit{carriage} or \textit{helicopter} but also in a scene different than the city roads in Cityscapes. Thus, while bigger OOD objects are easier to detect, the background objects might also be often detected as OOD. We observe this trend for results corresponding to prototype heatmaps in PROWL with STEGO. Although Maskomaly reports good results on SMIYC benchmark, we still observe that the masks detected for OOD are highly insufficient with only partial detections. In contrast, since CutLER first detects foreground object boxes and then provides semantic masks, the entire mask is precisely detected as OOD object using PROWL. For FS-Static, the OOD objects are synthetic images which often have a matching texture with the background of Cityscapes images. 
 Thus, we see some false predictions with the background pixels for Maskomaly and PROWL in the image of \textit{dog crossing the road}. The primary difficulty in this dataset could be missing the OOD object due to similar texture as background. This is apparent for the Maskomaly image with \textit{dog sitting on road}. 
 We observe that PROWL with CutLER most precisely detects the OOD masks in most cases. For the RO dataset, the OOD objects are meant as real obstacles of small sizes and varying numbers placed on the ego-track at different distances. Due to smaller sizes with increasing distance, all methods show worse performance as compared to other datasets.  The reproduced Maskomaly results falsely captures parts of the road. PROWL finds the OOD objects in most cases but also falsely predicts many background pixels as OOD. We observe the superior qualitative performance of object masks provided by STEGO and CutLER which helps in localising the objects as entire masks and then detecting them as OOD via prototype heatmaps in PROWL. We note although STEGO localises all the OOD object masks, the semantic masks are somewhat discontinuous. In comparison, PROWL with CutLER precisely localises and segments the OOD object masks, justifying the overall better performance. Although vanilla PROWL is quite close to detecting the OOD pixels, however with above examples we show that the additional unsupervised foreground masks helping in refining the OOD object masks avoiding spurious false predictions. 

To show generalisability to totally different domain and OOD objects, we further show impressive results on one of the most complex Indian Driving Dataset (IDD) \cite{varma2019idd} with totally different scene as compared to urban driving Cityscapes in Fig. 6 in Supplementary.

\vspace{-1mm}
\subsection{Plug-and-play application to other domains}
\vspace{-1mm}
\label{sec:results_other-odd}

\begin{figure}[htb]
    \vspace{-2mm}
    \centering
     \includegraphics[width=\linewidth]{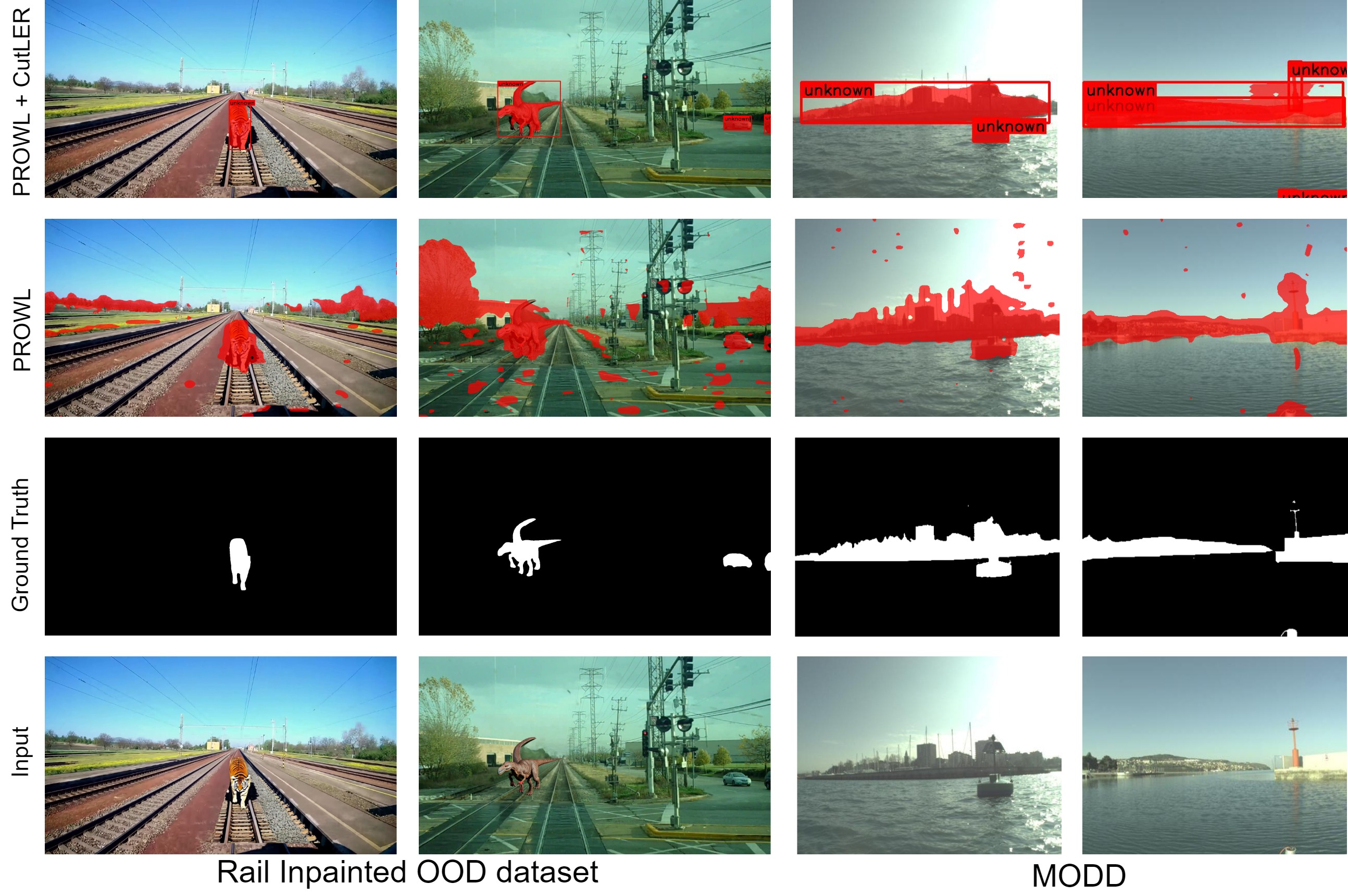}
   \vspace{-4mm}  
   \caption{Qualitative results for zero-shot OOD detection in other ODD domains - rail and maritime scene. Detected OOD pixel or segmentation masks are shown in red.
   }
   \label{fig:railsem_modd}
   \vspace{-3mm}
\end{figure}

\begin{table}[htbp]
\vspace{-1mm}
\centering
\resizebox{0.95\linewidth}{!}{%
\begin{tabular}{llllll}
\hline
                    &            & \multicolumn{2}{c}{Rail Inpainted OOD} & \multicolumn{2}{c}{MODD}        \\ \cline{3-6} 
Method              & Zero-shot & IoU               & F1                & IoU           & F1             \\ \hline
PROWL   & Yes         & 8.01               & 14.33             & 62.35          & 76.36          \\
PROWL + CutLER       & Yes         & \textbf{83.29}     & \textbf{90.38}    & \textbf{73.30} & \textbf{84.08} \\ \hline
\end{tabular}%
}
\caption{Performance of our zero-shot methods in other domains such as rail and maritime scenes for binary segmentation with fixed thresholds on the validation datasets.  
}
\label{tab:other_odd-thresh-fix}
\vspace{-4mm}
\end{table}

To demonstrate plug-and-play application of PROWL to other domains, we extend our evaluation to rail and maritime ODD scene with respective OOD datasets given in Sec. \ref{sec:datasets}. For rail scene, we use in-house created dataset with in-painted OOD objects (see Suppl.).
For simplicity, we consider the following $6$ predominant classes in RailSem19 \cite{zendelRailSem19DatasetSemantic2019} for creating prototype bank of ODD classes - \textit{train car, platform, rail, fence, person, pole}.  
Similarly, for maritime we refer to MODD \cite{bovconMaSTr1325DatasetTraining2019} and create prototype bank with \textit{sky, sea} as known ODD classes. Every other object is OOD. Since there exists no benchmark on RailSem19 OOD detection and the MODS \cite{bovconMODSUSVorientedObject2022} uses separate evaluation strategy, we provide evaluation using our zero-shot methods with PROWL. Since CutLER was trained only on unlabelled ImageNet \cite{dengImageNetLargescaleHierarchical2009}, it can easily provide zero-shot inference on any domain, whereas STEGO still needs to train on the datasets of new domain although without labels. Similarly, our supervised methods like Maskomaly \cite{ackermannMaskomalyZeroShotMask2023a}, RbA \cite{nayalRbASegmentingUnknown2023} relies on Mask2Former which still needs to train with the labels of the respective training data, thus can not be compared.

In Table \ref{tab:other_odd-thresh-fix}, we show the performance of PROWL vs PROWL with CutLER as a binary segmentation task using fixed default INCS threshold for detection of OOD pixels as provided in Sec. \ref{sec:exp_setup}. We find that PROWL with CutLER performs better as compared to vanilla PROWL in both cases. In Fig. \ref{fig:railsem_modd}, we show corresponding qualitative results. We observe that PROWL is significantly worse in case of Rail Inpainted OOD data while for MODD, PROWL with CutLER performs comparably slightly better than PROWL.This could be attributed to the fact that prototype heatmaps in PROWL provide per-pixel outputs, causing false positives to the background objects like \textit{vegetation, buildings} that are not in the assumed ODD class list. CutLER provides an advantage here, where the relevant object masks (including OOD) can be filtered based on foreground score while irrelevant background objects can be ignored. However, in MODD dataset, the task is relatively simple as everything other than \textit{sea, sky} should be detected as OOD or obstacle, thus both methods perform well on this dataset.

\vspace{-1mm}
\subsection{Ablation Study}
\vspace{-1mm}
\label{sec:results_ablation}

\begin{figure}[htbp]
    \vspace{-2mm}
    \centering
     \includegraphics[width=\linewidth]{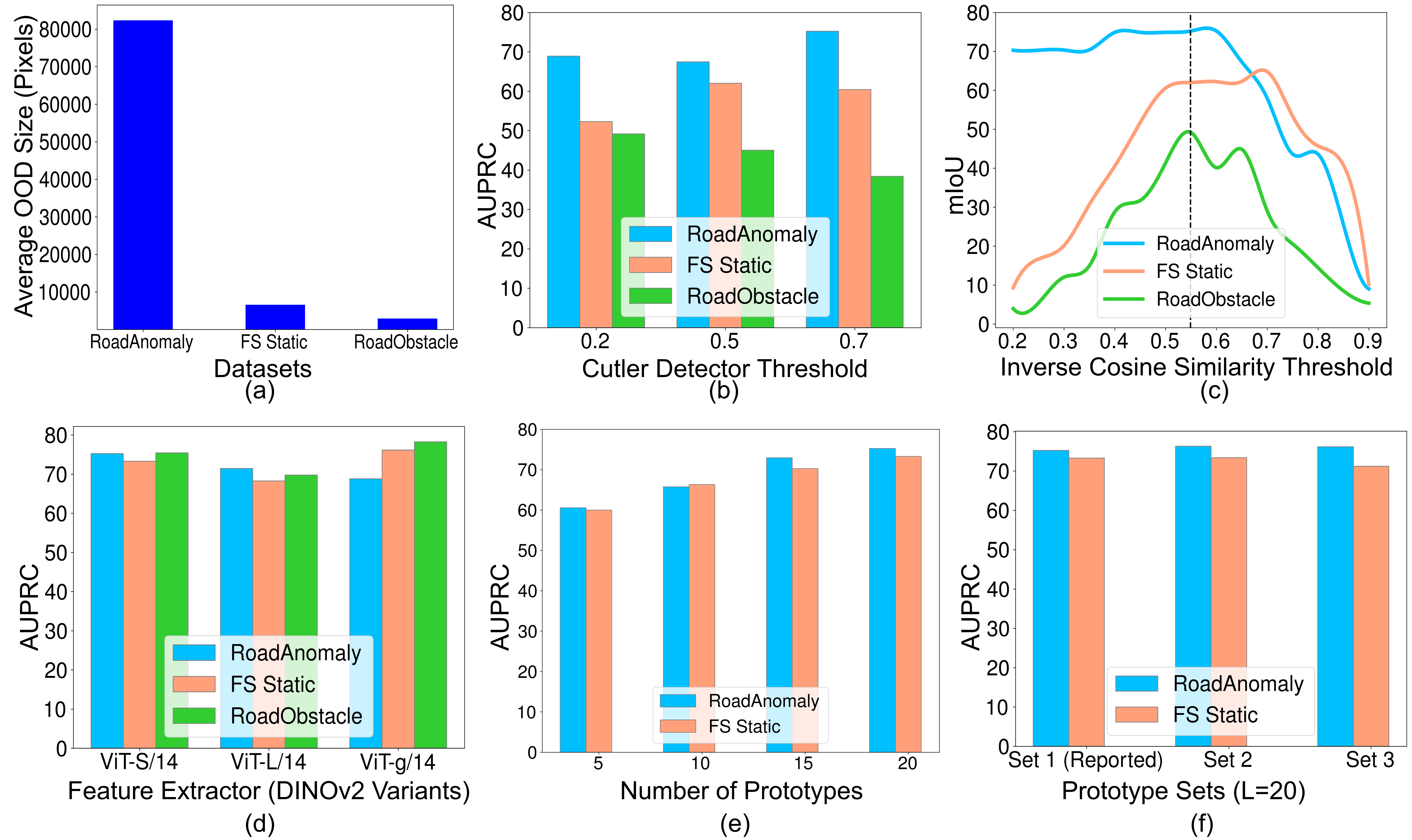}
   \caption{Ablation Study: a) Variation of OOD object size, and further studies on performance of PROWL  with variation of - b) Cutler Detector Threshold, c) Inverse Similarity Threshold, d) Feature Extractor models, e) Sets of $20$ prototypes, f) Number of prototypes used to create feature bank. All experiments conducted for different OOD datasets, given ODD classes from Cityscapes in road driving scene.
   }
   \label{fig:ablation_study}
   \vspace{-4mm}
\end{figure}

In Fig. \ref{fig:ablation_study}, we show the average OOD object size in pixels (a) as well as ablation studies for different hyperparameters used in the experiments with PROWL (b-f). We can see that RA contains much bigger objects compared to RO and FS Static. 
In (b), we investiagte the influence of different CutLER thresholds for generating optimal number of foreground masks for PROWL with the INCS threshold set at default value of $0.55$. Optimal mIoU is achieved for CutLER thresholds of $0.7$, $0.5$, and $0.2$ for RA, FS Static, and RO datasets respectively. We argue that smaller objects require lower thresholds to be detected reliably but also introduce more detection noise, whereas bigger objects are hard to miss even with a high detector threshold.
In (c), we fix the CutLER thresholds given above and find the optimum INCS threshold for best OOD detection based on IoU. We observe FS Static peaks at relatively high threshold of 0.7 as the dataset is quite similar to Cityscapes data, whereas RA and RO dataset peak at $0.55$, $0.60$ respectively close to the default threshold at $0.55$.
We also compare performance with different DINOv2 model variants as feature extractors in (d). Larger models like ViT-g/14 perform marginally better, however ViT-S/14 seems to offer a better trade-off considering significantly lower inference times.
Lastly, we analyze the influence of the prototype samples on the performance in terms of numbers (e) and selection (f). We show that already quite good AUPRC is achieved with as less as $5$ prototypes and it starts to saturate from $15$ and above. Furthermore, the choice of the specific prototype images seems to have less of an influence as all three distinct sets of samples tested lead to very similar performance. However, further investigation of this property is needed for samples derived from other datasets.

\vspace{-4mm}
\section{Conclusion}
\vspace{-2mm}

In this work, we proposed the first framework for zero-shot inference on vision foundation models for unsupervised OOD detection and segmentation - PROWL. It is a plug-and-play framework which can be easily transferred to different domains without further model training or fine-tuning on domain specific data. Since it relies on extracting prototype features from foundation models trained without labels, it stands as a practical approach towards open world settings. We show that PROWL combined with CutLER outperforms all the zero-shot as well supervised methods (trained without auxiliary OOD data) on RoadObstacle datasets and comparably with RoadAnomaly in road driving scenes. With different OOD datasets, we show that it can detect real OOD objects of different sizes (RoadObstacle) as well as diverse scenes beyond urban driving (RoadAnomaly, IDD). By applying it to rail and maritime applications, we demonstrate that it can be easily adapted to other domains. Even with a limited number of classes and prototypes defined in our ODD setting, PROWL performs reliably on the available benchmarks. However, due to the limited availability of diverse datasets with OOD objects for evaluation, a next step would be to put PROWL to the test in diverse scenarios with extensively defined ODD. Further, we clearly identified a need for harmonized evaluation metrics and benchmarks to enable a fair comparison of zero-shot approaches beyond metrics in SMIYC. 

\vspace{-4mm}
\section*{Acknowledgement}
\vspace{-2mm}

This work has been funded by the European Union and the German Federal Ministry for Economic Affairs and Climate Action as part of the safe.trAIn project.

{\small
\bibliographystyle{ieee_fullname}
\bibliography{main.bib}
}

\appendix
\section*{Supplementary}
\setcounter{section}{0} 

\begin{figure*}[htbp]
    \centering 
\begin{subfigure}{0.24\textwidth}
  \includegraphics[width=\linewidth]{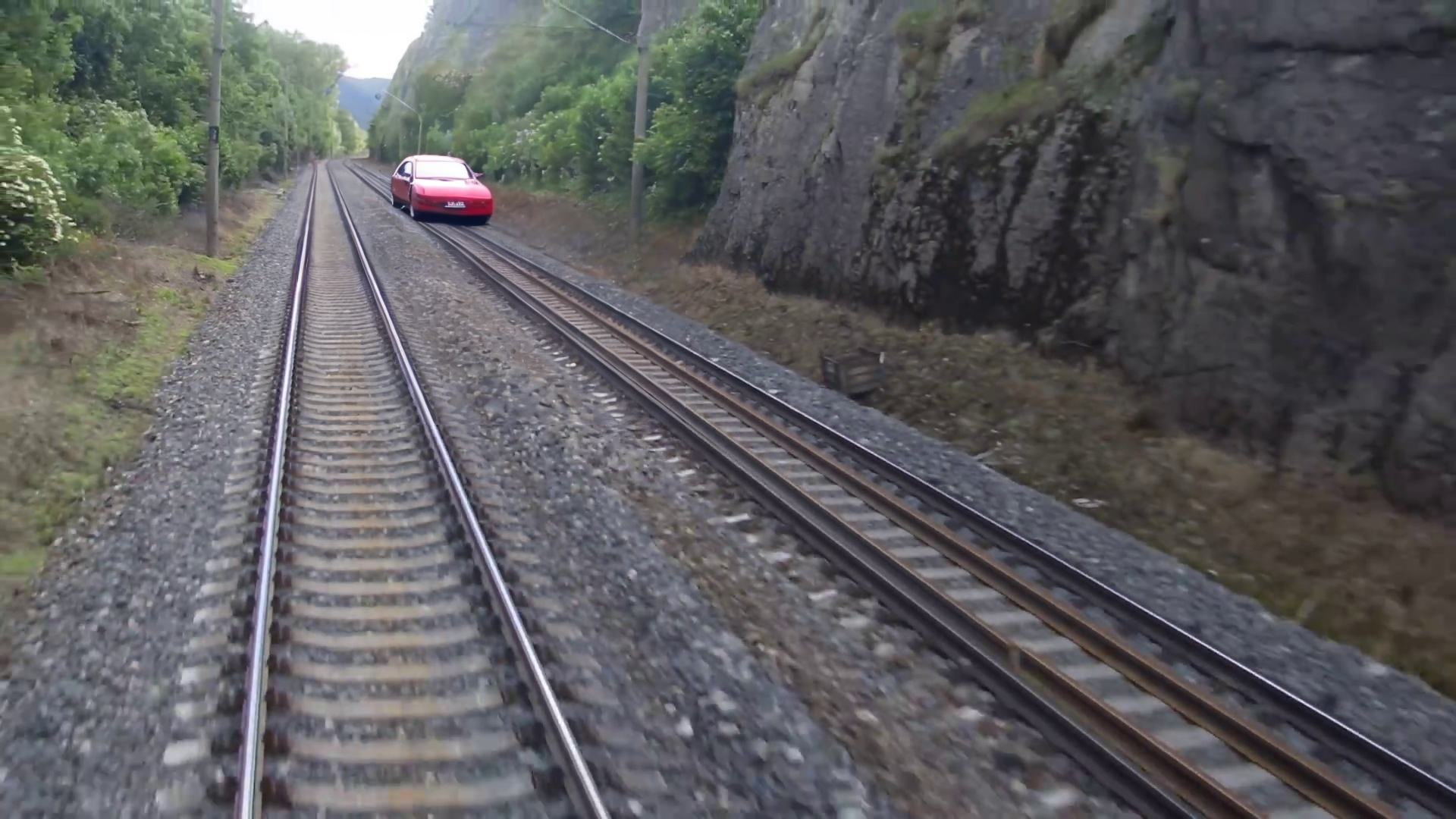}
\end{subfigure}\hfil 
\begin{subfigure}{0.24\textwidth}
  \includegraphics[width=\linewidth]{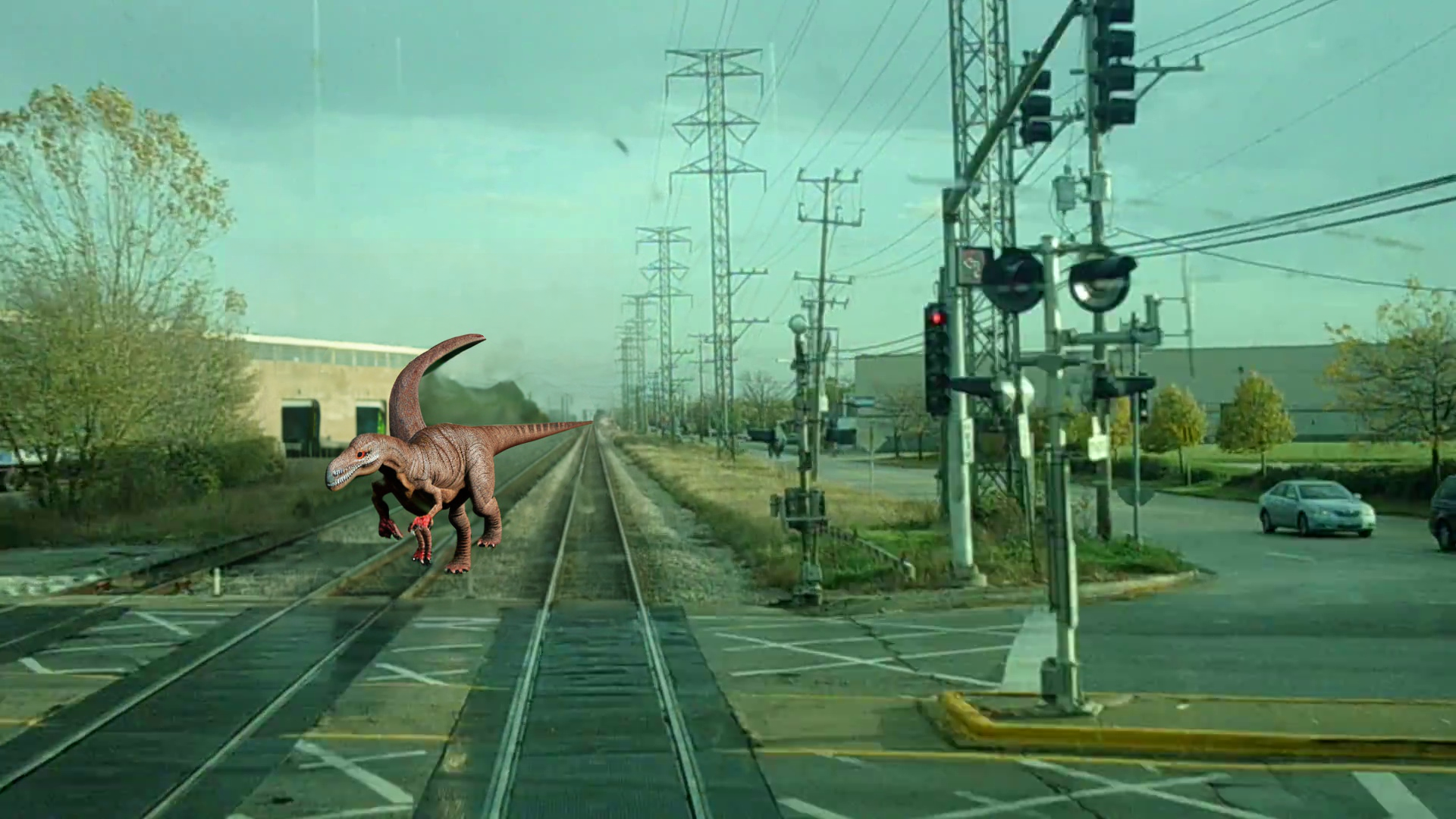}
\end{subfigure}\hfil 
\begin{subfigure}{0.24\textwidth}
  \includegraphics[width=\linewidth]{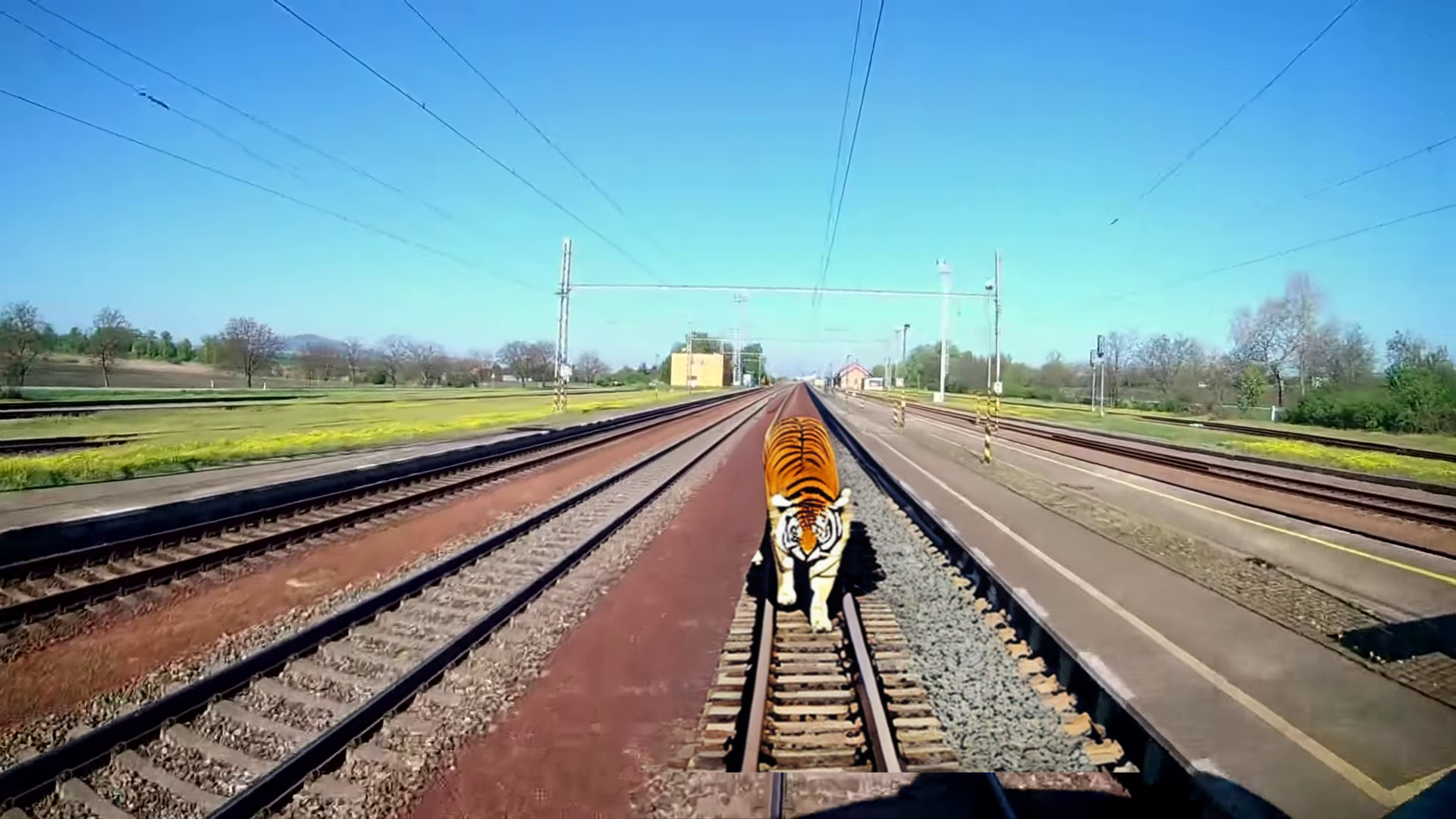}
\end{subfigure}\hfil
\begin{subfigure}{0.24\textwidth}
  \includegraphics[width=\linewidth]{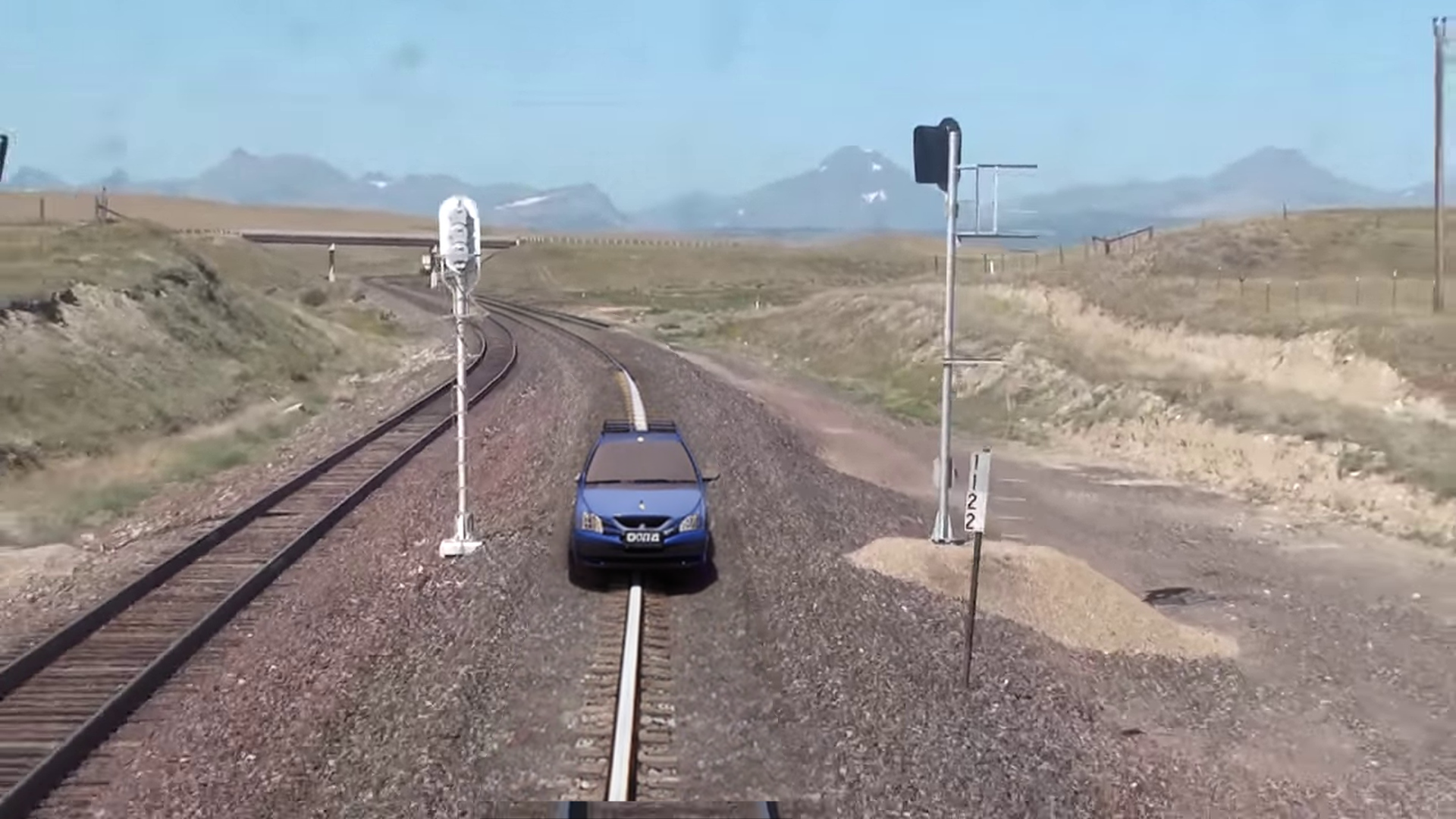}
\end{subfigure}

\begin{subfigure}{0.24\textwidth}
  \includegraphics[width=\linewidth]{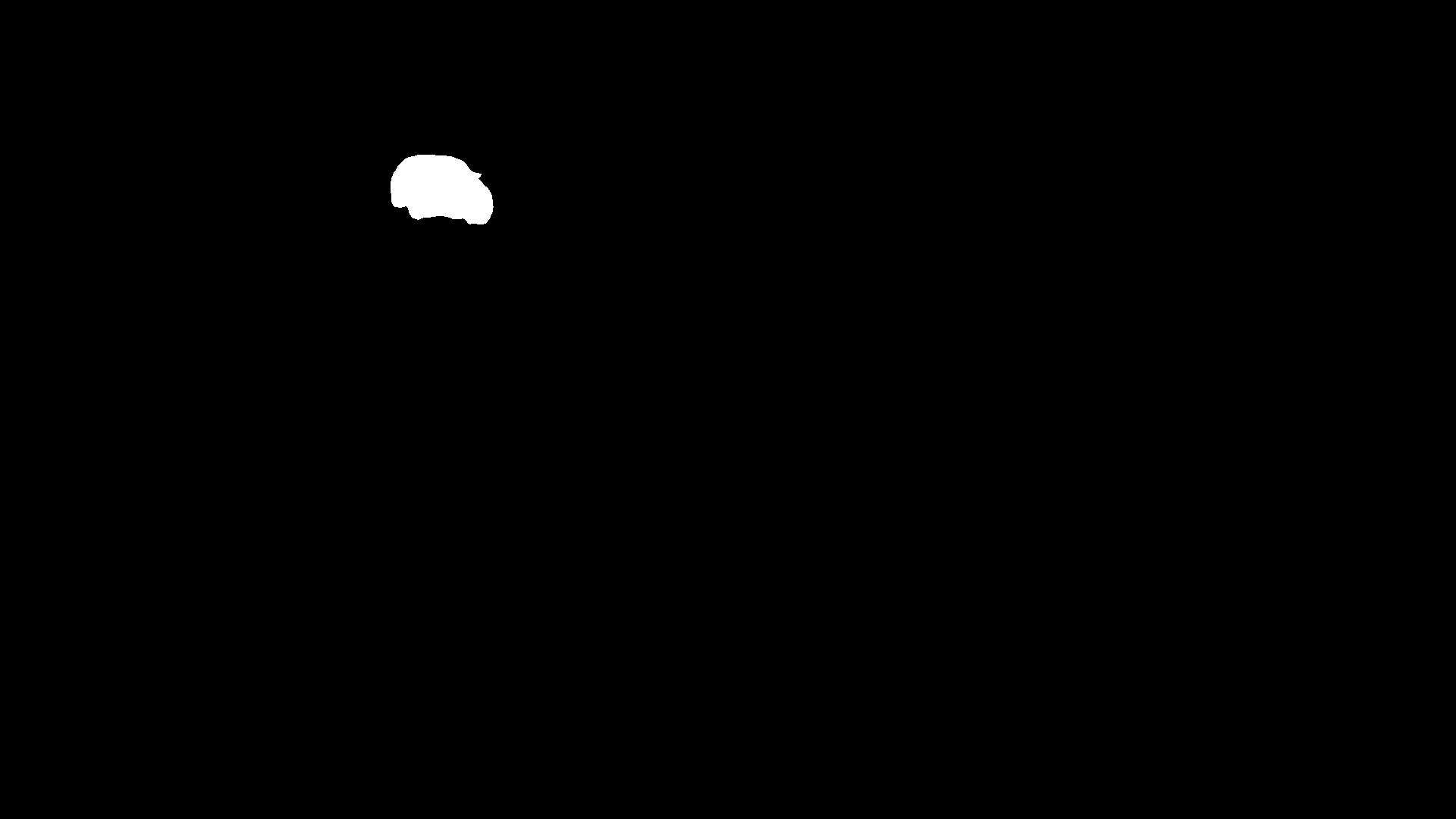}
\end{subfigure}\hfil 
\begin{subfigure}{0.24\textwidth}
  \includegraphics[width=\linewidth]{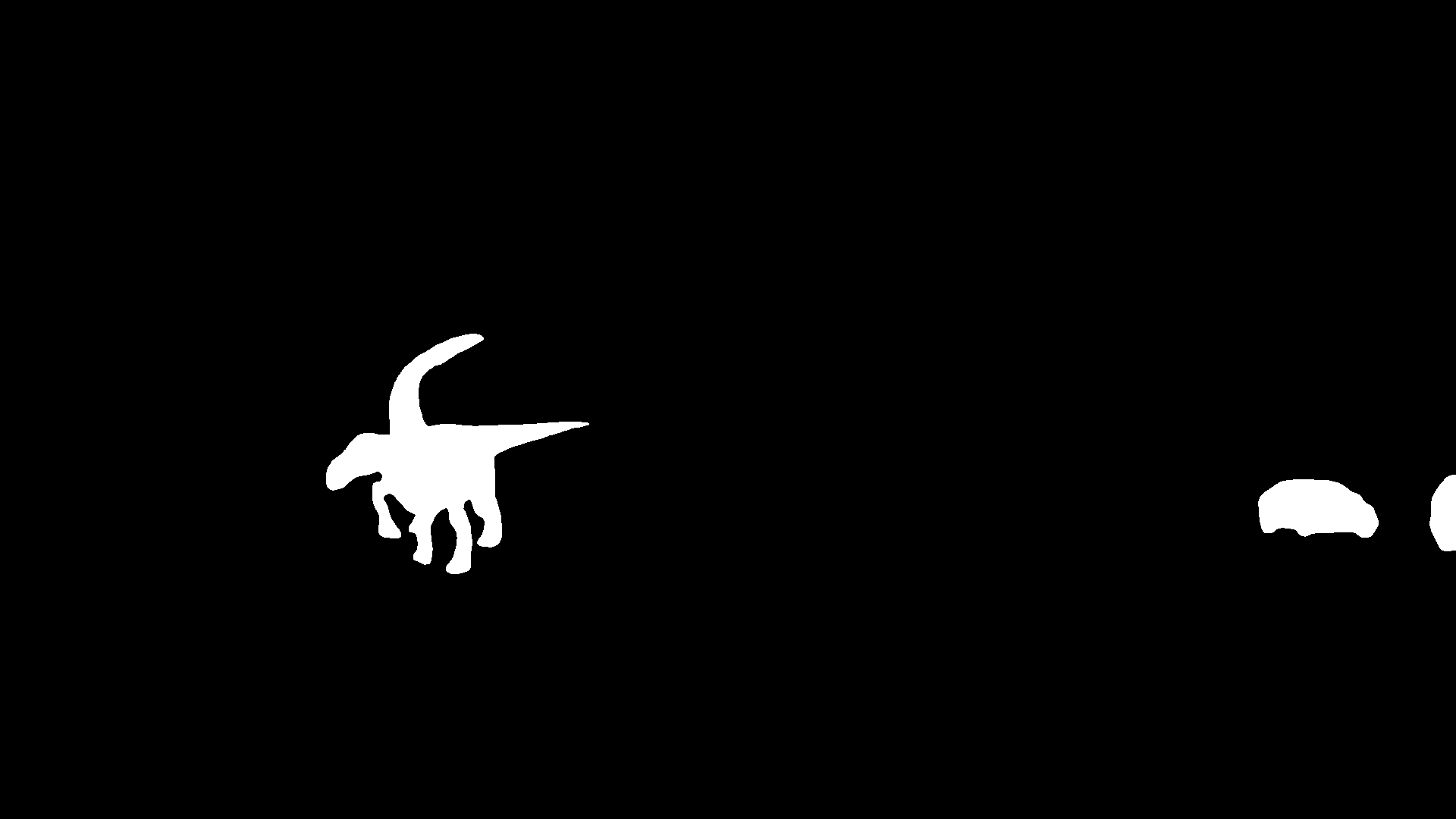}
\end{subfigure}\hfil 
\begin{subfigure}{0.24\textwidth}
  \includegraphics[width=\linewidth]{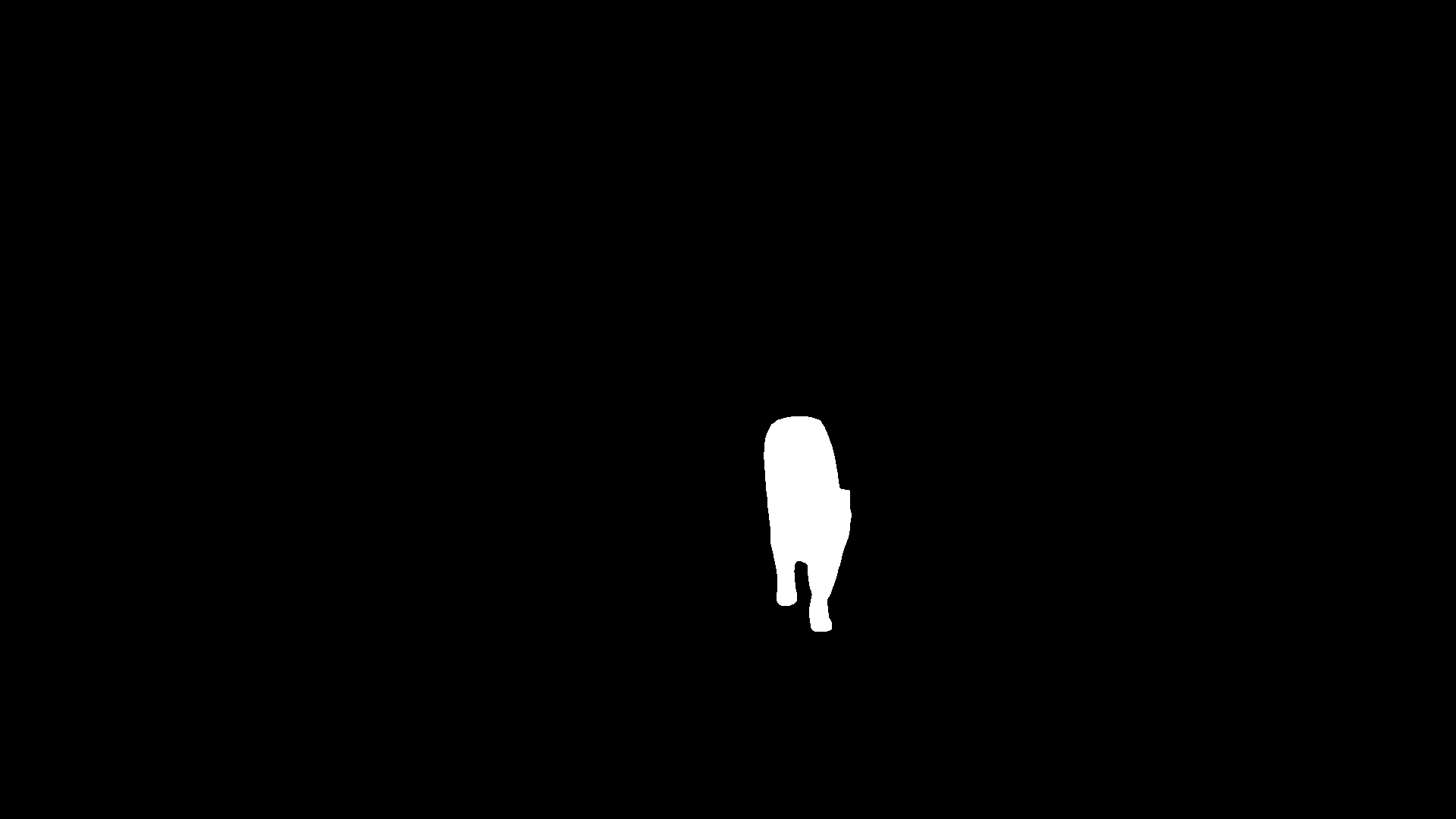}
\end{subfigure}\hfil
\begin{subfigure}{0.24\textwidth}
  \includegraphics[width=\linewidth]{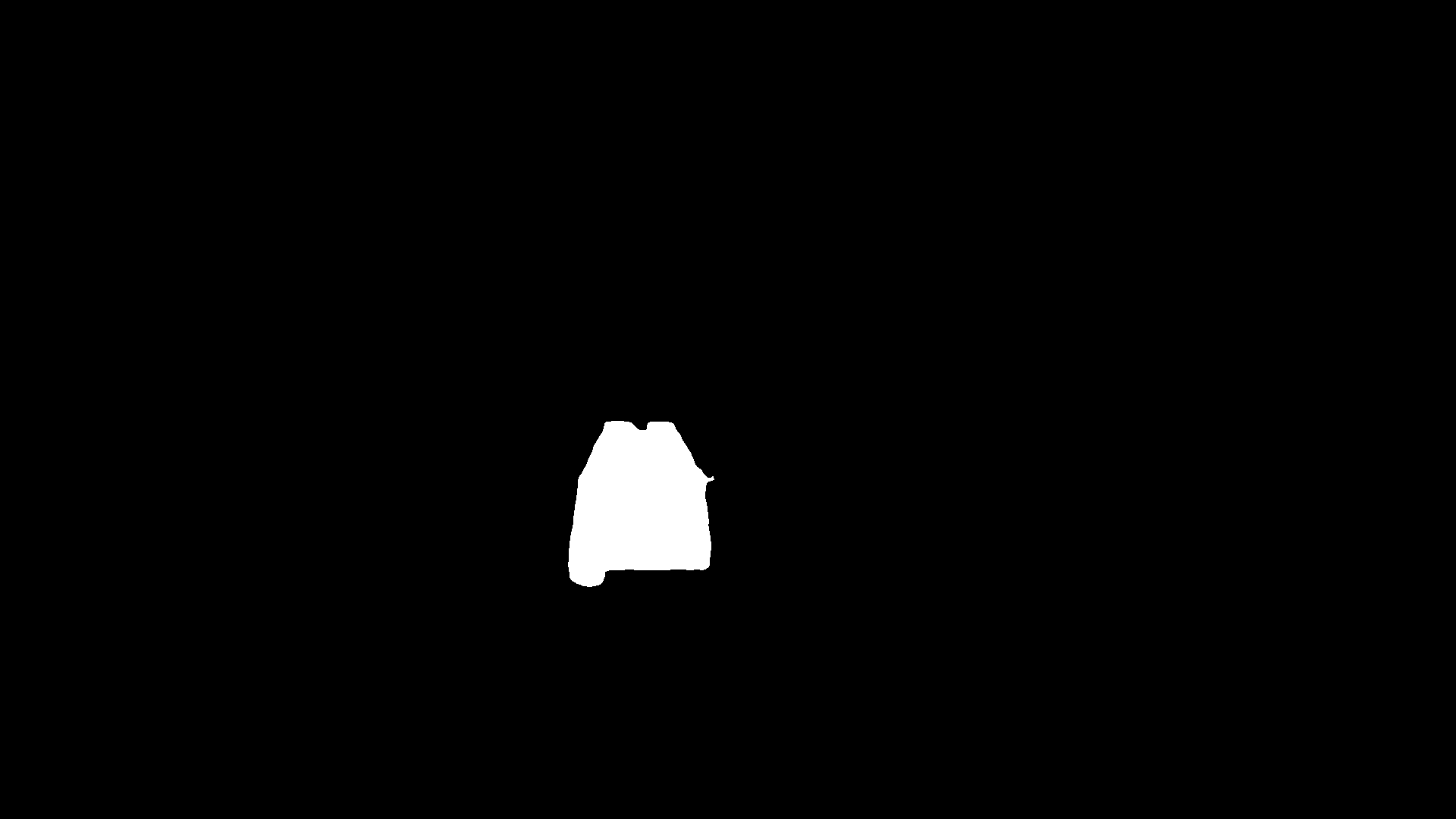}
\end{subfigure}
\caption{Sample images from OOD dataset created via inpainting OOD objects in RailSem19 dataset \cite{zendelRailSem19DatasetSemantic2019} (top-row) and the corresponding binary masks for the OOD objects (bottom-row).}
\label{fig:rs_images}
\end{figure*}

\section{Details about creating OOD dataset via Inpainting}

Due to absence of publicly available dataset in the rail ODD scene, we created an OOD dataset via inpainting OOD objects to the validation images of RailSem19 dataset \cite{zendelRailSem19DatasetSemantic2019}. Although generated images pertain to rail scene, but this method can be generically applied to any domain. The creation of the dataset involves relies on two methods - `Inpaint-Anything' \cite{yuInpaintAnythingSegment2023} and `Segment Anything Model' (SAM) \cite{kirillovSegmentAnything2023}. Inpaint-Anything takes some coordinates in an image and replaces the object which lies in the given coordinates. This object is replaced with the object that needs to be in-painted using text prompts, with the help of a diffusion model. Thus, the input image is changed to an image with the desired object given in the prompt at the specific location provided.

However, image generation using Inpaint-Anything is limited in the cases where there are no plausible objects to be replaced. Moreover, since the replaced object is in-painted, the corresponding OOD object mask is not obtained for utilising as the Ground-Truth (GT). To create the GT masks, we store the coordinate locations of the replaced object. Now, we leverage SAM by feeding the transformed image to it and also specifying the stored coordinate locations to generate the segmentation mask of the object associated with the coordinate locations. Thus, we create the image with the OOD object at the specified locations as well as the corresponding segmentation masks for the OOD objects in the image, as shown in Fig. \ref{fig:rs_images}.

\section{Additional Results}

In this section, we show further results and insights of the comparative performance of our method zero-shot method PROWL and its variants resulting from combination unsupervised segmentation methods, STEGO and CutLER. In Sec.
Firstly, we show the results for the segmentation outputs on the test set of In-Distribution (ID) datasets for each domain. Further, we show additional comparative results for OOD detection on the test set of the OOD datasets for Cityscapes \cite{cordtsCityscapesDatasetSemantic2016}.

\begin{figure*}[htb]
    \centering
    \includegraphics[width=0.9\linewidth]{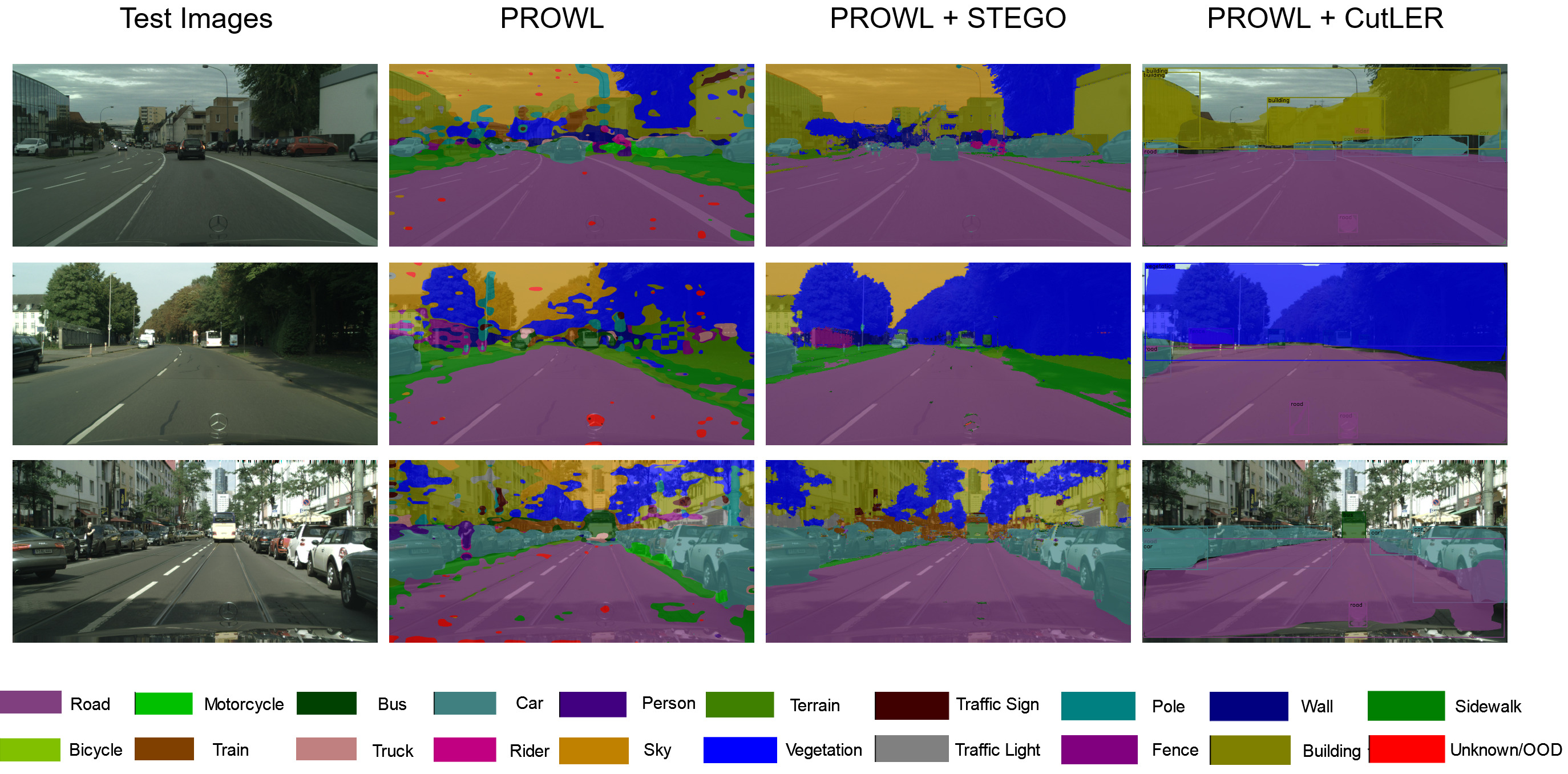}

    \caption{Performance comparison of our zero-shot methods on the segmentation outputs for the ODD classes on ID test dataset for road driving scene, i.e. Cityscapes \cite{cordtsCityscapesDatasetSemantic2016}.
   }
    \label{fig:sup_city}
    \vspace{-1em}
\end{figure*}

\subsection{Performance comparison on the ODD classes of ID test datasets}
\label{sec:supp_results_id}

\begin{figure*}[htb]
    \centering
    \includegraphics[width=0.85\linewidth]{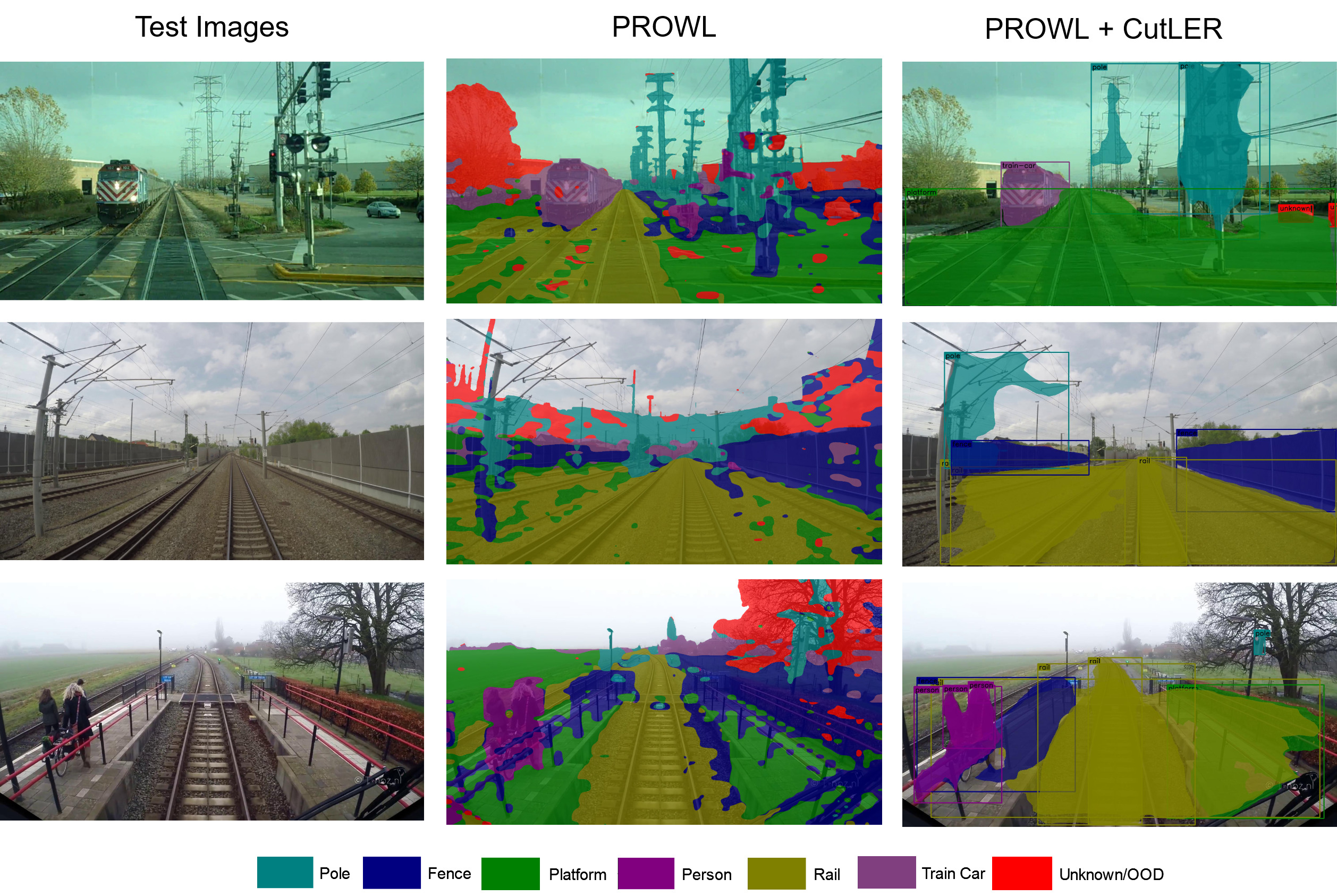}

    \caption{Performance comparison of our zero-shot methods on the segmentation outputs for the ODD classes on ID test dataset for rail scene, i.e. RailSem19 \cite{zendelRailSem19DatasetSemantic2019}.
   }
    \label{fig:sup_rail}
\end{figure*}

Here, we show the prototype-based segmentation outputs for the ODD classes for different variants of PROWL on the test set of the ID datasets used for each domain, i.e. Cityscapes for road driving scene (Fig. \ref{fig:sup_city}) and RailSem19 for rail scene (Fig. \ref{fig:sup_rail}). 

Fig. \ref{fig:sup_city} shows the segmentation from the test set images, for all $19$ ODD classes of Cityscapes  used to create the prototype feature bank. PROWL shows pixel-wise classification, whereas STEGO and CutLER provides semantic and instance segmentation masks combined with pixel-wise classification of PROWL. While both STEGO and CutLER generate unsupervised foreground masks, STEGO generates per-pixel output due to contrastive clustering of the ID train data whereas CutLER generates object boxes for foreground objects and then provides instance segmentation masks. Thus, CutLER provides segmentation for foreground object masks while ignoring background, like the \textit{sky} is ignored in all the three test images as well as the \textit{buildings} in last test image where they relatively lie in the background. Although, all these models have been trained for segmentation without labels, the overall segmentation is quite good. In PROWL, some noisy output is obtained (pixels in red shown in red) due to per-pixel classification based on ODD prototype classes. However, this is taken care of when combined with mask based evaluation using PROWL in STEGO and CutLER. \textit{We note the importance of having good quality prototype feature bank as this reflects the performance in correctly classifying ODD classes or OOD pixels.} For example, in the segmentation GT for \textit{road} in the prototype features include the test vehicle along with \textit{Mercedes logo} and thus they have been labeled as \textit{road}.

Fig. \ref{fig:sup_rail}, we show segmentation outputs test split of RailSem19, for PROWL and PROWL with CutLER for the assumed simple ODD list with $6$ classes - \textit{train car, platform, rail, fence, person, pole}. Since STEGO relies on unsupervised contrastive training on domain dataset and did not provide pre-trained model weights for RailSem19, we exclude it from comparison. We note although both zero-shot methods perform quite well on the ID test set, PROWL shows some OOD or unknown regions in red. This is primarily due to pixel-wise prototype matching where the pixels in red mostly correspond to classes like \textit{vegetation}, not defined in our current ODD list. In PROWL + CUTLER, we do not directly detect \textit{vegetation} as OOD as they are not detected as foreground masks and feature as background in the given test images since they are present over quite a distance. However, since \textit{car} is not defined in ODD list, it is detected as OOD in the first test image. Thus, sufficiently defining ODD class list is crucial while detecting OOD / unknown objects to avoid false predictions.

\begin{figure*}[htb]
    \centering
    \includegraphics[width=0.9\linewidth]{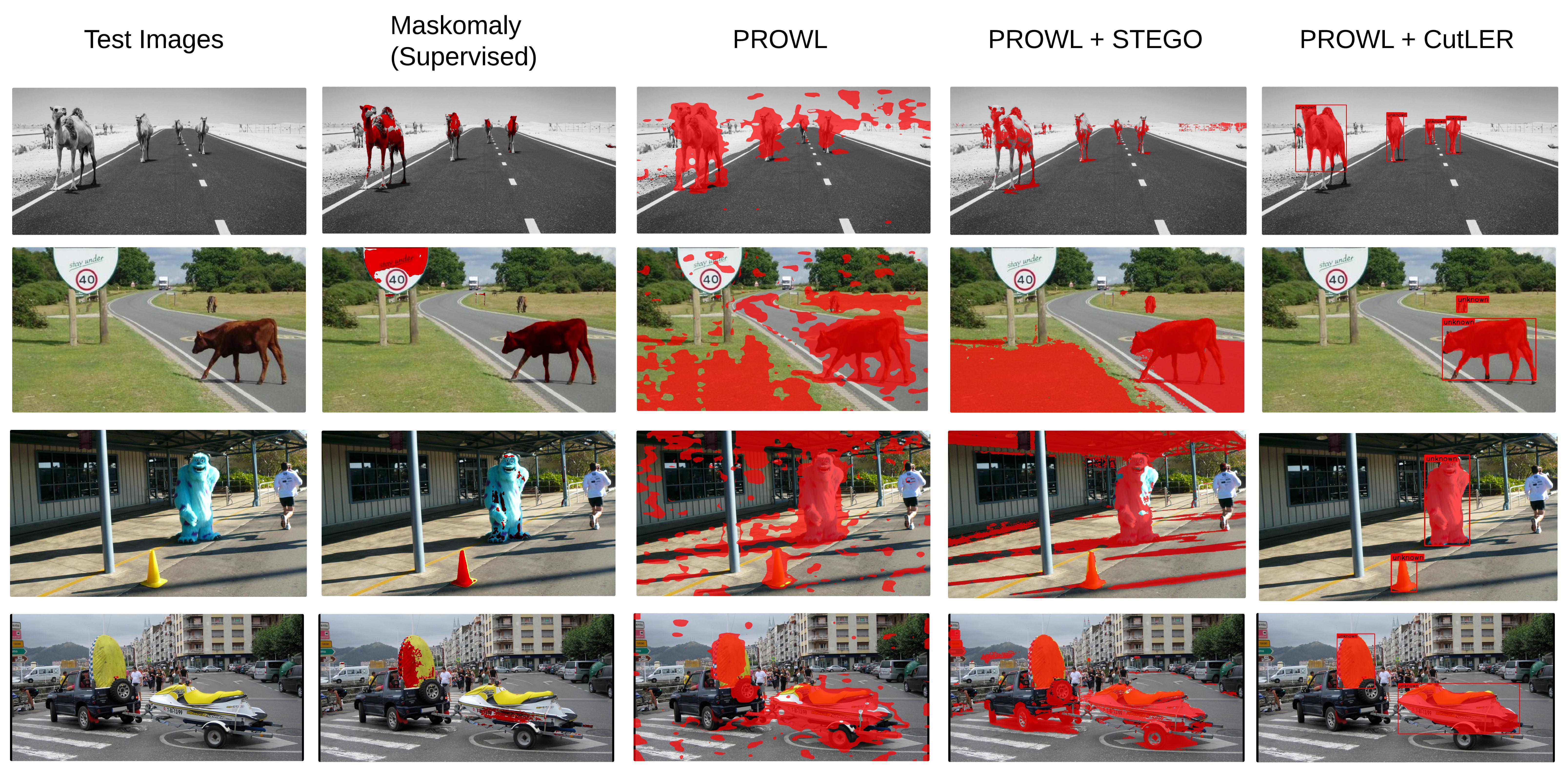}

    \caption{Performance comparison of our proposed zero-shot methods compared to supervised baseline for OOD object detection and segmentation on the test images of RoadAnomaly OOD (SMIYC-Anomaly Track) dataset. Detected OOD pixels are shown in red.
   }
    \label{fig:supp_anomaly_test}
\end{figure*}

\begin{figure*}[htb]
    \centering
    \includegraphics[width=0.9\linewidth]{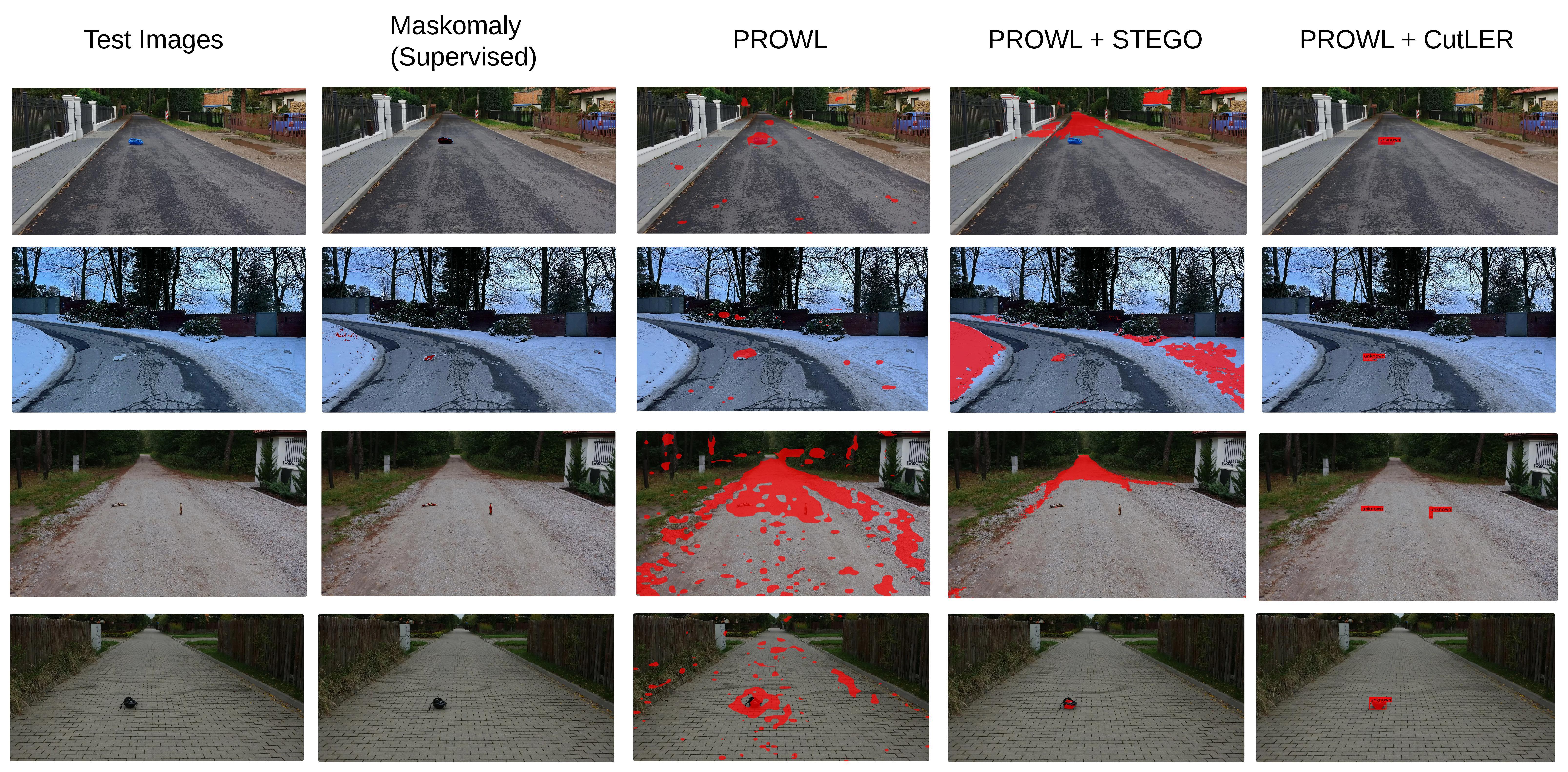}

    \caption{Performance comparison of our proposed zero-shot methods compared to supervised baseline for OOD object detection and segmentation on the test images of RoadObstacle (SMIYC- Obstacle Track) OOD dataset. Detected OOD pixels are shown in red.
   }
   
    \label{fig:supp_obstacle_test}
\end{figure*}

\begin{figure*}[htb]
    \centering
    \includegraphics[width=0.8\linewidth]{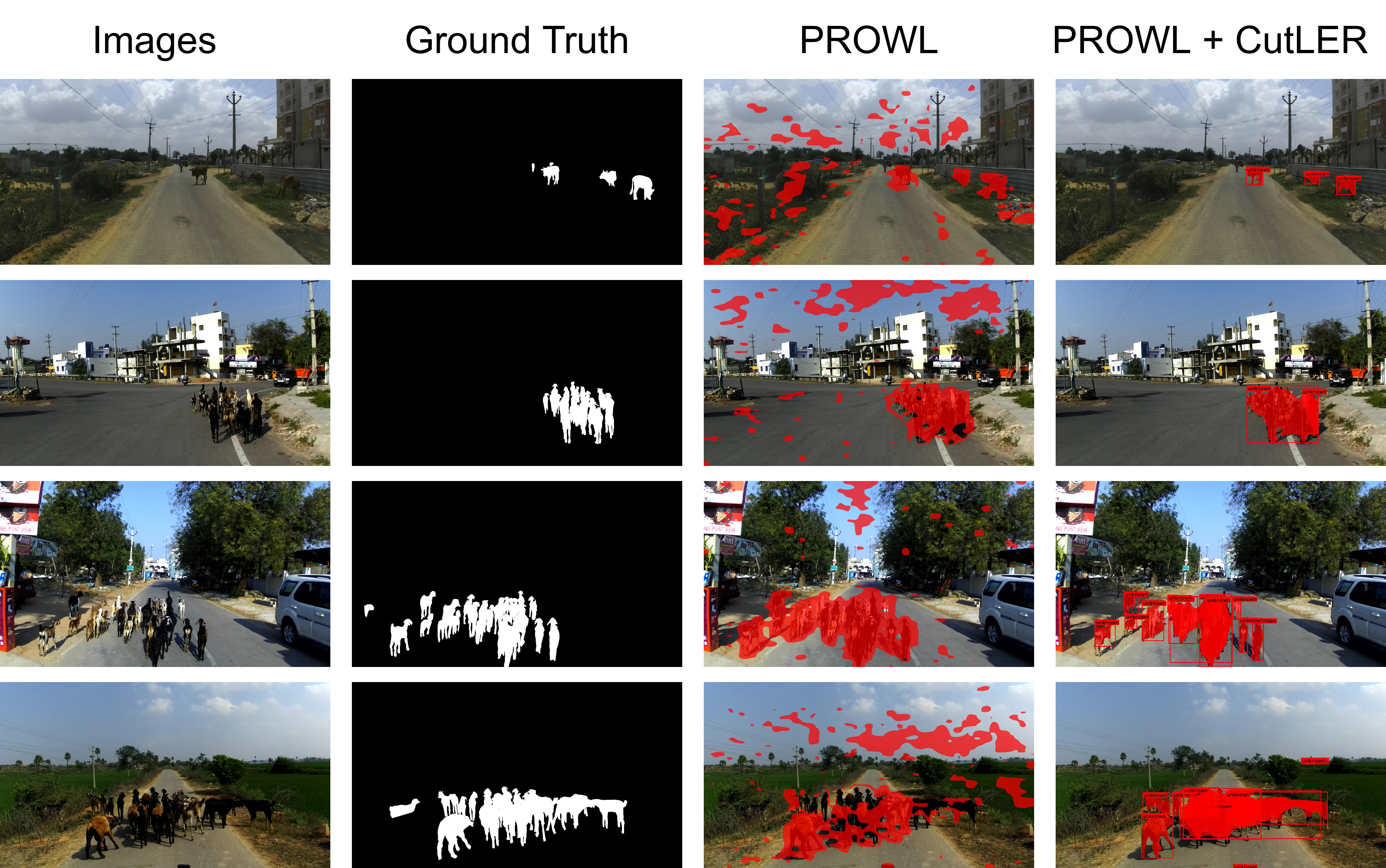}

    \caption{Qualitative performance comparison of our proposed zero-shot methods for OOD object detection and segmentation on the images from the Indian Driving Dataset\cite{varma2019idd}. Detected OOD pixels are shown in red.
   }
   
    \label{fig:IDD_OOD}
\end{figure*}

\subsection{Performance comparison on OOD test datasets}
\label{sec:supp_results_ood_test}

Here, we show additional results for OOD detection using our PROWL and it's variants compared to supervised baseline (Maskomaly \cite{ackermannMaskomalyZeroShotMask2023a}) particularly for the \textit{test} images for the given OOD datasets (RoadAnomaly and RoadObstacle) given in SMIYC benchmark \cite{chanSegmentMeIfYouCanBenchmarkAnomalya}, i.e Anomaly track and Obstacle track respectively, for Cityscapes \cite{cordtsCityscapesDatasetSemantic2016} as ID dataset. We show only qualitative results on the test sets due to absence of GT in the benchmark. For fair comparison, we use fixed confidence threshold of $0.9$ as suggested by authors of Maskomaly \cite{ackermannMaskomalyZeroShotMask2023a}. Similarly, for our methods - PROWL and its variants, we used inverse cosine similarity threshold fixed at $0.55$. GT segmentation masks for OOD objects for these test images are not provided, thus we only show qualitative results in Fig. \ref{fig:supp_anomaly_test} and \ref{fig:supp_obstacle_test} with detected OOD objects in red.

Fig. \ref{fig:supp_anomaly_test} show performance comparison on the RoadAnomaly dataset where the OOD objects are relatively bigger and the scenes are different than city road scenes in Cityscapes. We observe that supervised Maskomaly although localises the OOD object in some cases, but does not properly segment the object. In second test image it falsely predicts \textit{traffic sign }as OOD while in the third image, it misses the \textit{dressed-up bear} as OOD object. PROWL and PROWL + STEGO localises all the OOD objects, however provides noisy segmentation including background pixels. PROWL + CutLER shows overall best performance with correctly localising and segmenting all the OOD objects.

Fig. \ref{fig:supp_obstacle_test} show performance comparison on the RoadObstacle dataset where the OOD objects are varying in sizes as well as the scenes in the test data show different weather conditions and different road types such as \textit{dark asphalt, gravel, paved} and so on. This is the most challenging dataset where most methods have difficulty in spotting small OOD objects lying very far away in diverse scenes. We observe that supervised Maskomaly localises the OOD objects in the first two test images, however fails to detect them in the last two images. PROWL and PROWL + STEGO show noisy detections whereas PROWL with CutLER localises and segments all the instances of OOD objects quite well. 

Fig. \ref{fig:IDD_OOD} shows zero-shot performance of our methods on a subset of Indian Driving Dataset (IDD) \cite{varma2019idd}.
IDD can easily be deemed as one of the most difficult datasets for the autonomous driving scene understanding, due to extensive traffic, crowds, and, non-regular structures on the side of the roads like different types of buildings, banners, heaps etc. Also, the presence of uncommon obstacles such as animals coming into sudden proximity of the vehicles on the road are expected to be quite a domain shift as compared European urban driving dataset such as Cityscapes. 
Thus, this dataset is one of the most challenging datasets for evaluating the performance of a model for OOD detection and segmentation. Since OOD objects are not explicitly specified in this dataset, we create a small OOD test subset of $20$ samples containing object classes, such as \textit{animals} which do not overlap with Cityscapes domain classes. Thus, using the prototype feature bank based on Cityscapes, we evaluate the zero-shot performance of our methods using the generic threshold of $0.55$ for INCS and $0.2$ for CutLER without requiring to fine-tune any threshold on the datasets. PROWL shows its efficacy in determining the pixel regions where the OOD objects might be present. Moreover, when we incorporate the CutLER together with, we get a more accurate OOD localization which helps in robust OOD detection and segmentation. In all sample images showing multiple instances of \textit{animals on the road } are accurately segmented. The quantitative performance of PROWL shows an average IOU value of $26.46$, and F1 value of $39.84$ over the dataset, and PROWL+CutLER has an average IOU value of $55.99$, and F1 value of $67.47$ respectively. We note there are other fine-grained objects appearing in the scene which often get detected as OOD, although they are not deemed so nor they are present in Cityscapes ODD list.

We note that possible cases of failure often appear when the images are too dark and foreground objects in the images are not sufficiently visible.

Overall, we show that PROWL with CutLER can be readily used for plug-and-play zero-shot inference without further training or fine-tuning on the domain data, which works well for both instance segmentation on ID datasets as well as OOD detection on OOD datasets as an zero-shot method which performs comparably and also outperforms SOTA supervised methods for some OOD datasets.



\end{document}